\documentclass[runningheads]{llncs}

\usepackage{eccv}

\usepackage{eccvabbrv}

\usepackage{graphicx}
\usepackage{booktabs}
\usepackage[normalem]{ulem}
\usepackage{makecell}
\usepackage{tikz}
\usepackage[linesnumbered,ruled,vlined]{algorithm2e}

\usepackage{siunitx}
\usepackage{algorithmicx}
\usepackage{xcolor}
\definecolor{myred}{RGB}{200,30,45}

\usepackage{wrapfig}
\usepackage{tabularx}
\usepackage{subcaption}
\usepackage{float}
\usepackage{adjustbox}

\usepackage[accsupp]{axessibility}

\usepackage{hyperref}

\usepackage{orcidlink}

\newcommand{\boldfw}{\fontseries{b}\selectfont}

\begin{document}

\title{Beyond Accuracy: Benchmarking Cross-Task Consistency in Unified Multimodal Models}

\author{Weixing Wang\inst{1}\thanks{Equal contribution} \orcidlink{0009-0000-3231-4635}\and 
Liudvikas Zekas\inst{1,*} \and
Anton Hackl\inst{1,*}\orcidlink{0009-0000-4996-8934}\and
Constantin Alexander Auga\inst{1,*} \orcidlink{0009-0002-6664-8849} \and
Parisa Shahabinejad\inst{1} \and
Jona Otholt\inst{1} \and
Antonio Rueda-Toicen\inst{1} \and
Gerard de Melo\inst{1} \orcidlink{0000-0002-2930-2059}
}

\authorrunning{Wang et al.}

\institute{Hasso Plattner Institute / University of Potsdam, Potsdam, Germany \\
\email{\{weixing.wang, gerard.demelo\}@hpi.de} \\
\email{\{Constantin.Auga, Anton.Hackl, Liudvikas.Zekas\}@student.hpi.uni-potsdam.de}
}

\maketitle

\begin{abstract}
Unified Multimodal Models (uMMs) aim to support both visual understanding and visual generation within a shared representation. However, existing evaluation protocols assess these two capabilities independently and do not examine whether they are semantically aligned. As a result, it remains unclear whether current uMMs learn coherent unified representations that remain consistent across tasks given a visual concept. We introduce \textbf{XTC-Bench}, a scene-graph-grounded evaluation framework that measures cross-task visual semantic consistency. By deriving both generation prompts and understanding queries from a structured scene graph, our framework enables fact-level alignment analysis across objects, attributes, and relations. We propose Continuous Cross-Task Agreement (CCTA), a fine-grained metric that quantifies semantic agreement between generation and understanding over matched atomic facts, isolating internal consistency from standalone task accuracy. Extensive experiments on eight open-source and one commercial unified models reveal that high generation or understanding performance does not imply strong cross-task alignment, and architectural analysis shows consistency is governed by how tightly learning objectives are coupled across modalities, not by architectural unification alone. XTC-Bench provides a reproducible and model-agnostic framework for diagnosing representation-level misalignment, offering a concrete direction for advancing unified multimodal modeling beyond isolated task performance.

\textit{Project page: \url{https://weixingw.github.io/xtc-bench/}}

\end{abstract}

\section{Introduction}
\label{sec:intro}

Unified Multimodal Models (uMMs) are designed to jointly support visual understanding and visual generation within a shared architectural and representational space~\cite{deng2025emergingpropertiesunifiedmultimodal, chen2025janusprounifiedmultimodalunderstanding, yang2025mmadamultimodallargediffusion, wu2025omnigen2explorationadvancedmultimodal, xie2025showosingletransformerunify, xie2025showo2improvednativeunified, han2025visiondialectunifyingvisual, ma2025unitokunifiedtokenizervisual}. By aligning vision and language into a shared latent representation, these models aim to enable seamless reasoning and synthesis across modalities. Recent systems demonstrate strong performance on both visual question answering and text-to-image generation benchmarks, suggesting substantial progress toward unified multimodal intelligence.

\begin{figure}[tb]
    \centering
    \includegraphics[width=0.90\linewidth,keepaspectratio]{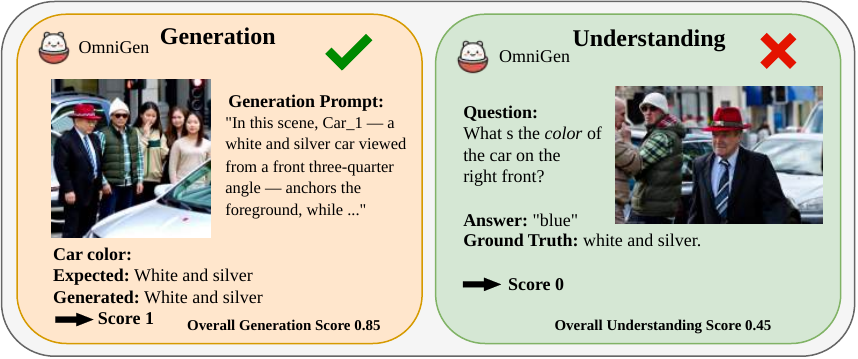}
    \vspace{0.3mm}
    \includegraphics[width=0.90\linewidth,keepaspectratio]{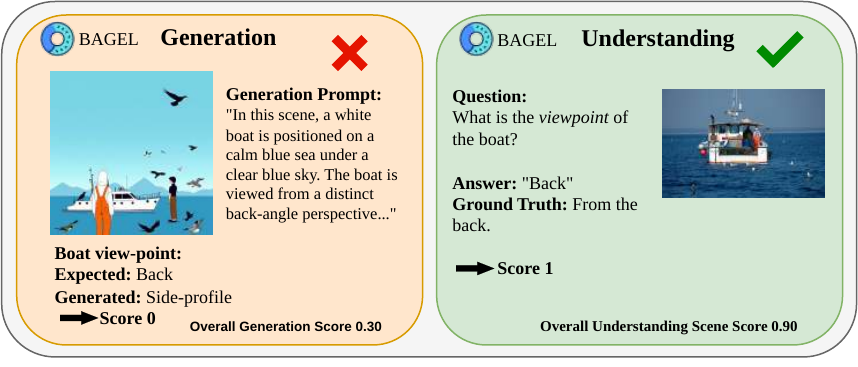}
    \caption{OmniGen can generate attribute details while failing to recognize them (top), whereas BAGEL understands relational viewpoints that it fails to render (bottom).}
    \label{fig:my_figure}
\end{figure}

However, sharing representation does not automatically imply semantic coherence across tasks. If a model truly learns a unified representation, then accessing this representation through perception (understanding) and through synthesis (generation) should yield mutually consistent semantic structures. In practice, a model may correctly answer fact-based questions while failing to render the same facts during generation, or generate correct visual details yet failing to recognize them during comprehension, as shown in \cref{fig:my_figure}. Such discrepancies indicate structural asymmetry in the internal representation.

Existing evaluation paradigms are not designed to expose this asymmetry. Visual understanding is typically assessed through VQA-style benchmarks~\cite{agrawal2016vqavisualquestionanswering,liu2024mmbenchmultimodalmodelallaround}, while generation quality is measured via alignment or perceptual metrics~\cite{ghosh2023genevalobjectfocusedframeworkevaluating,huang2025t2icompbenchenhancedcomprehensivebenchmark,chong2020effectivelyunbiasedfidinception}. Although recent unified benchmarks broaden modality coverage~\cite{wang2025xmodbenchbenchmarkingcrossmodalcapabilities,zou2026unimmmumassivemultidisciplinemultimodal,shi2025realunifyunifiedmodelstruly,li2025unievalunifiedholisticevaluation,xie2025mmeunifycomprehensivebenchmarkunified}, they primarily report performance per task without grounding both directions in the same structured semantic representation. Consequently, high understanding and generation scores may conceal internal semantic misalignment that remains undetected.

To address this limitation, we introduce XTC-Bench, a scene-graph-grounded benchmark for evaluating cross-task visual semantic consistency. Scene graphs provide an explicit, structured representation of objects, attributes, and relations within a scene. In XTC-Bench, a reference scene graph serves as a unified semantic anchor from which both generation prompts and fact-aligned VQA queries are deterministically derived. The generated image is parsed into a predicted scene graph using the same extraction pipeline, enabling direct fact-level comparison between what a model generates and what it understands.

\begin{figure}[tb]
  \centering
  \includegraphics[width=\linewidth]{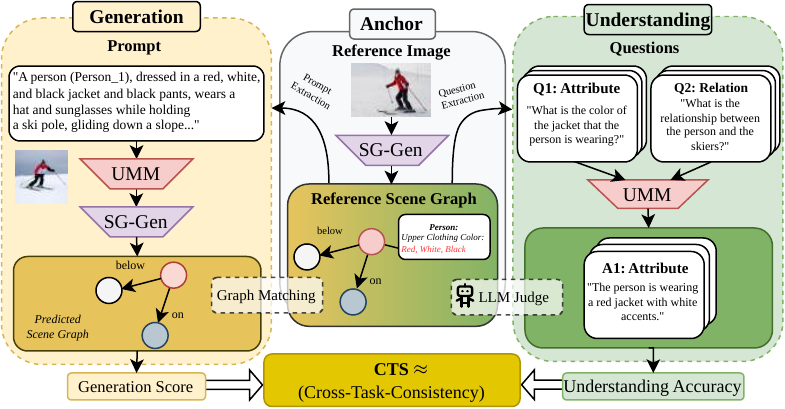}
  \caption{Overview of our evaluation framework. Cross-Task Consistency is evaluated by aggregating Understanding and Generation performance anchored on the same scene.}
  \label{fig:full_pipeline}
\end{figure}

We evaluate models along three complementary dimensions: generation quality (G), understanding performance (U), and cross-task consistency measured by Continuous Cross-Task Agreement (CCTA). This design disentangles structural coverage, factual correctness, and internal alignment, isolating representation-level coherence from raw task accuracy. \cref{fig:full_pipeline} depicts the high-level concept of our evaluation framework. 

Across eight open-source unified models and two commercial systems, we observe a systematic pattern: generation and understanding performance are only weakly predictive of cross-task consistency. Better performance in one direction does not reliably translate to improved alignment. Attribute grounding and relational reasoning exhibit particularly pronounced discrepancies, revealing that unified architectures often achieve functional integration without synchronized semantic representations. These findings suggest that evaluating unified multimodal models requires moving beyond isolated accuracy metrics toward explicit measurement of internal cross-task agreement. XTC-Bench provides a reproducible and model-agnostic framework for diagnosing representation-level misalignment and offers a principled foundation for advancing unified multimodal modeling. Our contributions are threefold:

\begin{enumerate}
    \item We formalize cross-task visual semantic consistency as a measurable property of unified multimodal models and introduce Continuous Cross-Task Agreement (CCTA) for fine-grained fact-level alignment analysis.
    \item We present XTC-Bench, a scene-graph-grounded and fully automated benchmark that jointly evaluates generation, understanding, and their semantic agreement under a shared structured representation.
    \item Through extensive empirical evaluation, we demonstrate that current uMMs exhibit systematic structural asymmetry between generation and understanding, motivating future research toward representation-level coherence.
\end{enumerate}

\section{Related Work}

\subsection{Unified Multimodal Models}

Unified Multimodal Models aim to construct general-purpose systems capable of both visual understanding and generation within a shared architecture. To realize this goal, different models explore distinct design strategies. Chameleon \cite{team2024chameleon} adopts an early-fusion, token-based transformer that handles text and image interchangeably within a shared discrete token space. Emu3 \cite{wang2024emu3} shows that pure autoregressive modeling can achieve competitive multimodal performance without relying on diffusion. Janus-Pro \cite{chen2025janusprounifiedmultimodalunderstanding} decouples vision encoders into separate pathways for understanding and generation. Show-o \cite{xie2025showosingletransformerunify} combines autoregressive and discrete diffusion modeling, using next-token prediction for text and masked-token prediction for images. BLIP3-o \cite{chen2025blip3ofamilyfullyopen} follows an LLM+diffusion architecture with sequential pretraining. BAGEL \cite{deng2025emergingpropertiesunifiedmultimodal} is trained on interleaved multimodal tokens, exhibiting emergent compositional reasoning capabilities. Despite these advances, it remains unclear whether a shared representation truly yields semantic coherence across tasks or whether current models merely combine both capabilities at an engineering level, which motivates the cross-task consistency evaluation proposed in this work.

\subsection{Unified Multimodal Model Benchmarks}
Evaluation benchmarks in multimodal AI have traditionally targeted either understanding or generation in isolation. Understanding benchmarks such as MME \cite{fu2025mmecomprehensiveevaluationbenchmark}, MMBench~\cite{liu2024mmbenchmultimodalmodelallaround}, and SEED-Bench~\cite{li2023seedbenchbenchmarkingmultimodalllms} assess perception and reasoning through VQA, multiple-choice, and captioning tasks. Generation benchmarks such as GenEval~\cite{ghosh2023genevalobjectfocusedframeworkevaluating} and T2I-CompBench++~\cite{huang2025t2icompbenchenhancedcomprehensivebenchmark} measure compositional prompt alignment alongside perceptual metrics such as FID~\cite{chong2020effectivelyunbiasedfidinception} and CLIP-based scores.

Several recent studies attempt more unified evaluations, yet each falls short of measuring semantic agreement between tasks. MME-Unify~\cite{xie2025mmeunifycomprehensivebenchmarkunified} and UniEval~\cite{li2025unievalunifiedholisticevaluation} provide holistic assessments across both tasks but still report per-task performance without measuring agreement between the two directions. RealUnify~\cite{shi2025realunifyunifiedmodelstruly} investigates whether one capability enhances the other, rather than whether both are grounded in a shared consistent representation. XModBench~\cite{wang2025xmodbenchbenchmarkingcrossmodalcapabilities} evaluates cross-modal consistency across modality permutations but remains confined to single-task reasoning formats. FysicsWorld~\cite{jiang2025fysicsworldunifiedfullmodalitybenchmark} extends coverage to any-to-any modality mappings yet prioritizes modality interaction over cross-task alignment under identical structured semantics. GAPEval~\cite{wang2026quantifyinggapunderstandinggeneration} introduces bidirectional formulations to quantify the performance gap between understanding and generation, but measures disparity rather than fine-grained fact-level agreement. No existing benchmark explicitly grounds both tasks in the same structured semantic representation and evaluates their agreement at
the level of atomic visual facts, which is the central contribution of this work.

\subsection{Scene Graphs}

Scene graphs provide a structured, fine-grained representation of image content by decomposing scenes into objects with their attributes, and directed edges representing pairwise relations between them \cite{chang2021comprehensive}. Scene graphs can be constructed from images via object detectors paired with relationship classifiers \cite{zellers2018neural, xu2017scene, dai2017detecting, lorenz2024fairpsgg}. We repurpose scene graphs as a unified representation for evaluating both visual understanding and generation, and for measuring their alignment. By breaking down visual content into key elements (e.g., objects, their attributes, and relationships), scene graphs enable precise, element-level comparison that global similarity scores such as CLIPScore \cite{hessel2021clipscore} cannot provide.

\section{XTC-Bench: Benchmarking Cross-Task Consistency}

We introduce XTC-Bench, a scene graph-grounded benchmark designed to evaluate cross-task semantic consistency in unified Multimodal Models.
Given a shared visual concept, XTC-Bench automatically measures whether a model’s understanding and generation behaviors are semantically aligned under a common structured representation.

\subsection{Cross-Task Consistency}

Let $I \in \mathcal{I}$ denote an image and $G = (\mathcal{V}, \mathcal{E})$ its corresponding scene graph, where each node $v_i \in \mathcal{V}$ represents an object with attribute set $\mathcal{A}(v_i)$ and each edge $e_{ij} \in \mathcal{E}$ encodes a pairwise semantic relation. The scene graph $G$ serves as the unified semantic anchor from which both tasks are derived:
a generation prompt $P_G$ verbalizes all objects, attributes, and relations in $G$, and a set of question-answer pairs $\mathcal{Q} = \{(q_k, y_k)\}_{k=1}^{K}$ is constructed such that each question corresponds to a single atomic fact in $G$.

Given a uMM $\mathcal{M}$, the generation branch produces an image $\hat{I} = \mathcal{M}_{\mathrm{gen}}(P_G)$, from which a predicted scene graph $\hat{G}$ is extracted using the same parsing pipeline applied to the reference image. The understanding branch independently answers each question as $\hat{y}_k = \mathcal{M}_{\mathrm{und}}(I, q_k)$. For each atomic fact $f \in G$, we denote the model's generation response as $g_f$ (reflected in $\hat{G}$) and its understanding response as $u_f$ (reflected in the VQA answers). Cross-task consistency is defined as the degree of agreement between $g_f$ and $u_f$.

This definition deliberately decouples consistency from accuracy. A model that excels at each task independently may be inconsistent if its generation and understanding behaviors contradict each other. By grounding both tasks in the same structured scene graph, XTC-Bench enables fact-level alignment analysis and extends evaluation toward representation-level coherence. \subsection{Dataset Construction Pipeline}

\begin{table}[h]
\centering
\caption{Dataset statistics of XTC-Bench.}
\label{tab:dataset_stats}
\resizebox{\columnwidth}{!}{
\begin{tabular}{l|cccccccc}
\toprule
\textbf{Dataset} & \makecell[c]{\textbf{Total}\\\textbf{Images}} & \makecell[c]{\textbf{Total}\\\textbf{Facts ($|F|$)}} & \makecell[c]{\textbf{Avg.}\\\textbf{Obj/Img}} & \makecell[c]{\textbf{Avg.}\\\textbf{Rel/Img}} & \makecell[c]{\textbf{Avg.}\\\textbf{Attr/Img}} & \makecell[c]{\textbf{\% Obj.}\\\textbf{Retr.}} & \makecell[c]{\textbf{\% Attr.}\\\textbf{Query}} & \makecell[c]{\textbf{\% Rel.}\\\textbf{Query}} \\
\midrule
COCO & 1,000 & 16,618 & 10.48 & 16.84 & 8.03 & 15.9\% & 39.5\% & 44.6\% \\
Visual Genome & 1,000 & 14,912 & \hphantom{0}8.22 & 11.70 & 6.76 & 16.1\% & 39.3\% & 44.6\% \\
\bottomrule
\end{tabular}
}
\end{table}

We construct XTC-Bench from 2,000 manually curated images sampled from the COCO 2017 validation set~\cite{lin2015microsoftcococommonobjects} and Visual Genome~\cite{krishna2016visualgenomeconnectinglanguage}. For each image, we generate a structured scene graph, a descriptive prompt, and a set of fact-aligned VQA samples. As detailed in \cref{tab:dataset_stats}, XTC-Bench contains over 31,000 unique facts ($|F|$) and exhibits high semantic density, averaging 8 to 10 objects and 11 to 17 relations per image. To comprehensively evaluate compositional understanding, the VQA queries are systematically distributed across relations (44.6\%), attributes (39.4\%), and object retrieval (16.1\%).

\subsubsection{Scene Graph Generation}
\label{sec:scene_graph_generation}
Scene graphs can be constructed using one-step joint prediction methods or two-step pipelines separating object detection and relation prediction.
We adopt the two-step Fair-PSG~\cite{lorenz2024fairpsgg} pipeline for improved modularity and control. First, object instances and segmentation masks are obtained using kMaX-DeepLab~\cite{yu2023kmaxdeeplabkmeansmasktransformer}. Let $R$ be the set of 57 possible predicates that can be predicted by Fair-PSG. One special predicate $\text{NR} \in R$ is the ``No-Relation'' predicate, which indicates the non-existence of a valid relation between a pair of objects $(i,j)$. Fair-PSG predicts a score $s_{ij}^{r}$ for each $r \in R$ as an initial ranking of the most relevant predicates between each pair of objects. We provide a detailed ablation study for the method selection and the full list of class names in the appendix. 

\subsubsection{Relation Refinement and Graph Merging}

Raw relation predictions favor recall, whereas XTC-Bench requires high precision to avoid false semantic supervision.
Let $s_{ij}^{\text{NR}}$ denote the predicted score of the ``No-Relation'' class between objects $i$ and $j$.
We filter candidate relations by thresholding at $s_{ij}^{\text{NR}}=0.5$. Additionally, for each object pair that passes the no-relation filter, only predicate classes $r$ with predicted score $s_{ij}^{r} \geq 0.4$ are retained.
The remaining candidate relations are then verified by Qwen3-VL-235B~\cite{Qwen3-VL} through a visual question answering step: the VLM receives the image annotated with bounding boxes and is asked to confirm or reject each candidate predicate with a structured Yes/No answer.

For exclusive predicates such as \textit{eating} and \textit{driving}, we enforce a uniqueness constraint: if multiple subjects share the same predicate toward one object, only the highest-scoring subject--object pair is kept.

During early experiments, we found that dense clusters of overlapping instances often produced scene graphs that were unnecessarily complex and difficult to interpret. To improve graph clarity, we merge overlapping instances of the same class into a single meta-node. Let $\mathcal{C}$ denote a group of instances whose padded bounding boxes overlap. We initially considered merging groups of size 2 or greater, but found that pairs of instances could still usually be described adequately without grouping. We therefore apply merging only when the group size is at least 3. After merging, $\mathcal{C}$ is replaced by a single meta-node and all intra-group relations are discarded.

\begin{figure}[tb]
  \centering
  \includegraphics[width=\linewidth,trim=0 0 0 0,clip]{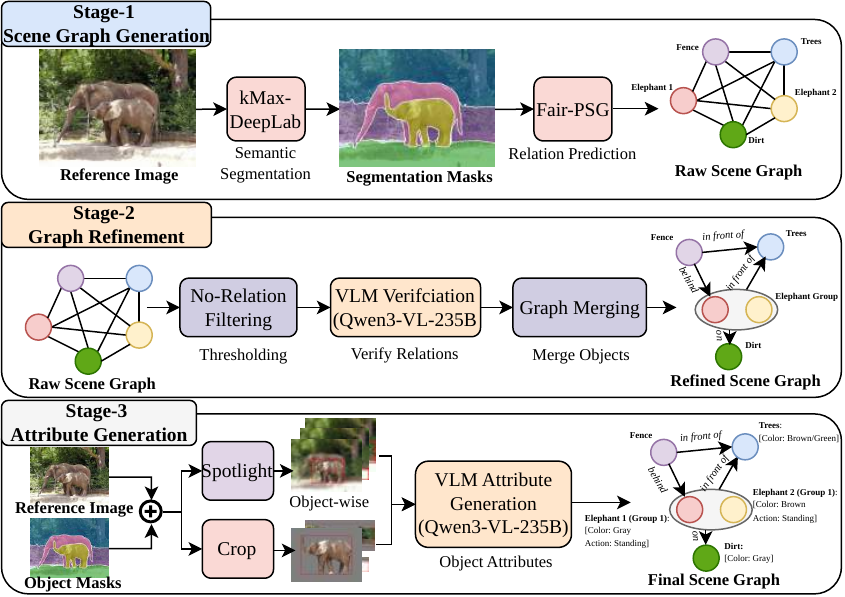}
  \caption{Scene-graph extraction pipeline, producing the final reference scene graph used for evaluation.}
  \label{fig:sg_gen_pipeline}
\end{figure}

\subsubsection{Attribute Generation}

To enrich object-level semantic descriptions, we define a set of meta-classes 
$\mathcal{C} = \{c_1, c_2, \dots, c_{|\mathcal{C}|}\}$ 
over object categories.
Each meta-class $c \in \mathcal{C}$ is associated with a predefined attribute key set $\mathcal{K}_c$.
Concretely, the meta-classes include 30 categories, including \textit{person}, \textit{vehicle}, \textit{infrastructure}, \etc. 
Each meta-class is equipped with semantically salient attribute keys such as color, material, or texture.

First experiments revealed two recurring failure modes. First, unconstrained attribute generation produced unstable and low-confidence outputs for some keys, especially material-related descriptions, with cross-run inconsistency (for example, the same bed being described as "rough" in one run and "smooth fabric" in another, or metallic objects being labeled “wooden”). To improve reproducibility, we constrained prediction to meta-class-specific key spaces and retained only attributes that can be visually grounded with high confidence. Second, object localization based on a simple crop or a full image with bounding boxes frequently caused target confusion. In scenes with a dominant foreground object and weak-background targets (for example, clouds), the model occasionally described the foreground instance instead of the intended target. Bounding-box-only prompting also failed for small instances and spatially diffuse objects, where the box covered large irrelevant regions.

For each object instance \(v_i\) assigned to meta-class \(c_j\), we therefore provide two images: (1) a \emph{Semantic Spotlight} view, where the full scene is shown but non-target regions are blurred while the target remains sharp with a bounding-box overlay, and (2) a \emph{Detail Crop}, where the target is isolated on a neutral gray background. This dual-view setup improves both identity disambiguation and fine-grained attribute recognition. We apply the same protocol across Qwen3-VL-7B, Qwen3-VL-32B, and Qwen3-VL-235B.

We apply structured prompting with visual chain-of-thought reasoning to predict attribute-value pairs within the predefined key space $\mathcal{K}_{c_j}$.
Formally, the attribute set of object $v_i$ is defined as
\begin{align}
\mathcal{A}(v_i)
=
\left\{
(k, a_{i,k})
\;\middle|\;
k \in \mathcal{K}_{c_j},\;
a_{i,k} = f_{\text{VLM}}(v_i^{\text{ROI}}, I, k)
\right\},
\end{align}
where $f_{\text{VLM}}$ denotes the inference function using Qwen3-VL-235B.

\subsubsection{Prompt Generation}

Given a scene graph $G$, we construct a linearization function $P_G = \mathcal{L}(G)$, which enumerates objects, attributes, and relations.
The initial linearization preserves all graph facts without introducing external information.
Subsequently, two LLM-based refinement steps are applied.
We deliberately split this process up into two stages to mitigate attribute entanglement and hallucination, which we frequently observed when large language models attempt to naturalize densely populated graphs in a single pass.
First, per-object refinement utilizes Qwen3-235B to enrich and naturalize each object's description by prompting the model to generate a concise caption based on the instance's extracted mask and the concrete attributes saved in its scene graph.
Isolating the refinement at the instance level ensures accurate attribute binding and prevents cross-contamination from neighboring objects.
Second, sentence-level refinement smooths the overall prompt for natural language flow, weaving the already-grounded descriptions into a cohesive sentence.

\subsubsection{VQA Generation}

For each atomic fact $f \in G$, we generate a corresponding question-answer pair.
Fact types include \textit{attribute queries} (``What is the color of the red car?''), \textit{object retrieval} (``What object is dark red and parked?''), and \textit{relation queries} (``What is the spatial position of the person relative to the bench?'').
Each question maps to exactly one fact to enable fine-grained evaluation.

When multiple object instances of the same class exist in the scene, we employ a combinatorial search over the target object's attributes to ensure unique identification. Specifically, the algorithm incrementally evaluates combinations of valid attributes of increasing size ($k=1,2,\dots$) until it identifies the minimal subset of values that do not overlap with those of competing instances. To robustly detect such overlaps, attribute values are tokenized and normalized into sets of lowercase tokens; an attribute combination is considered ambiguous if any set intersection or direct substring match occurs with a competing object. This iterative deepening guarantees that the target object is uniquely grounded using the shortest possible description.

For relation queries, parallel edges representing multiple relationships between the same subject-object pair are aggregated into a single comprehensive question to prevent redundant evaluation.

\subsubsection{Graph Matching}

To compare $G$ and $\hat{G}$, we perform a two-stage semantic graph matching process, summarized in Algorithm~\ref{alg:graph_matching}. Nodes are first grouped by label and aligned using the Hungarian algorithm. For each label group present in both graphs, we build a cost matrix $C$ between ground-truth nodes $v_i$ and predicted nodes $\hat{v}_j$ that combines attribute similarity ($\text{sim}_{\text{attr}}$) and incident edge structural similarity ($\text{sim}_{\text{edge}}$):
\begin{equation}
    C_{ij} = \alpha \cdot (1 - \text{sim}_{\text{attr}}(v_i, \hat{v}_j)) + \beta \cdot (1 - \text{sim}_{\text{edge}}(v_i, \hat{v}_j)).
\end{equation}
Here, $\text{sim}_{\text{attr}}$ calculates attribute similarity using Sentence Transformer embeddings (\eg, the \texttt{all-MiniLM-L6-v2} checkpoint~\cite{reimers-2019-sentence-bert}). Specifically, for each individual attribute key $k$ (e.g., color, material) present in the nodes, we sort and concatenate its values into a single string $s_k$, which is embedded into a vector $\mathbf{h}_k = \text{enc}(s_k)$. We then calculate pairwise cosine similarities for individual attributes: $\cos(\mathbf{h}_{i,k}, \mathbf{h}_{j,k})$. If an attribute key exists in one node but not the other, it is penalized with a similarity of 0. The overall $\text{sim}_{\text{attr}}(v_i, \hat{v}_j)$ is the arithmetic mean of these pairwise similarities across the union of all attribute keys from both nodes.

The edge structural similarity $\text{sim}_{\text{edge}}$ measures local topological agreement with a sub-level Hungarian matching that aligns the edges incident to $v_i$ with those incident to $\hat{v}_j$. The sub-level cost for a pair of incident edges $(e, \hat{e})$ is defined as $1 - \text{sim}_{\text{val}}(e, \hat{e})$, where $\text{sim}_{\text{val}}$ incorporates both the textual embedding of the relations and the attributes of the connected neighbor nodes $u$ and $\hat{u}$:
\begin{equation}
    \text{sim}_{\text{val}}(e, \hat{e}) = 0.4 \cdot \cos(\mathbf{z}_e, \mathbf{z}_{\hat{e}}) + 0.6 \cdot \frac{\text{sim}_{\text{attr}}(u, \hat{u}) + \text{sim}_{\text{attr}}(v_i, \hat{v}_j)}{2}.
\end{equation}
Here, $\mathbf{z}_e$ is the spatial sentence embedding of the concatenated relation text ``\texttt{[source\_label] [relation] [target\_label]}''. 

By setting $\alpha=0.7$ and $\beta=0.3$, we enforce a higher priority on attribute matching. We arrived at this parameter configuration by comparing attribute dominance ($\alpha=0.7$ and $\beta=0.3$), similarity equality ($\alpha=0.5$ and $\beta=0.5$), and relation dominance ($\alpha=0.3$ and $\beta=0.7$). We found that the resulting node alignments differed across the three configurations in approximately $5\%$ of the matched nodes. By evaluating a sample of these discordant assignments between the ground-truth and predicted scene graphs, we determined that attribute dominance yielded the most accurate matches.

We solve for the globally optimal assignment $\pi^* = \arg\min_{\pi} \sum_{i} C_{i,\pi(i)}$ restricted to matching nodes of the exact same label. We do not use a maximum cost rejection threshold ($C_{\max}$) for the alignment; the Hungarian algorithm simply matches nodes sharing the same label until one set is exhausted. Any remaining nodes due to a mismatched count are then discarded. 

\begin{algorithm}[t]
\caption{Two-Stage Semantic Graph Matching}
\label{alg:graph_matching}
\SetAlgoLined
\KwIn{Ground-truth graph $G$, Predicted graph $\hat{G}$}
\KwOut{Matched node pairs $\mathcal{M}_{\text{nodes}}$, Matched edge pairs $\mathcal{M}_{\text{edges}}$}
$\mathcal{M}_{\text{nodes}} \leftarrow \emptyset$, $\mathcal{M}_{\text{edges}} \leftarrow \emptyset$\;
Group all nodes in $G$ and $\hat{G}$ by exact class labels $L$\;
\ForEach{label $l \in L$ present in both graphs}{
    Let $V_l \subseteq G$ and $\hat{V}_l \subseteq \hat{G}$ be nodes with label $l$\;
    Initialize cost matrix $C$ of size $|V_l| \times |\hat{V}_l|$\;
    \ForEach{node pair $(v_i \in V_l, \hat{v}_j \in \hat{V}_l)$}{
        Compute $\text{sim}_{\text{attr}}(v_i, \hat{v}_j)$ via mean per-key attribute embedding similarity\;
        Compute $\text{sim}_{\text{edge}}(v_i, \hat{v}_j)$ via sub-level Hungarian incident edge matching\;
        $C_{ij} \leftarrow \alpha \cdot (1 - \text{sim}_{\text{attr}}) + \beta \cdot (1 - \text{sim}_{\text{edge}})$\;
    }
    Solve optimal assignment $\pi^*$ on $C$ using Hungarian algorithm\;
    Add matched pairs $(v_i, \hat{v}_{\pi^*(i)})$ to $\mathcal{M}_{\text{nodes}}$\;
}
\ForEach{ground-truth edge $e = (u, v) \in G$}{
    \If{both $u$ and $v$ are matched in $\mathcal{M}_{\text{nodes}}$ to $(\hat{u}, \hat{v})$}{
        \If{edge $\hat{e}$ exists between $\hat{u}$ and $\hat{v}$ in $\hat{G}$}{
            Add $(e, \hat{e})$ to $\mathcal{M}_{\text{edges}}$\;
        }
    }
}
\Return{$\mathcal{M}_{\text{nodes}}$, $\mathcal{M}_{\text{edges}}$}
\end{algorithm}

Finally, edge matching is performed as a secondary step after the node matching. Because there can be at most one edge between two nodes, we match the edges if they are present between aligned node pairs $(v_i, \hat{v}_j)$, and we do not match them if they are not present.

\subsection{Evaluation Metrics}

We evaluate models along two axes: \textbf{Generation} and \textbf{Understanding}, 
and measure their consistency via \textbf{Continuous Cross-Task Agreement (CCTA)}.

\subsubsection{Generation Score}

Instead of exact string matching, we adopt an LLM-as-judge protocol for generation evaluation that is better aligned with human preferences~\cite{2024improvingautomaticvqaevaluation,gu2025surveyllmasajudge,ging2024openendedvqabenchmarkingvisionlanguage}.
We convert structured predictions (objects, attributes, relations, and graph matching) into VQA-style verification questions and let an LLM assign a score $g_f \in [0,5]$ for each atomic fact $f$.

Specifically, for each VQA question generated from the ground-truth scene graph, we look up the corresponding answer from the predicted scene graph using the node mapping established by graph matching, and let the LLM judge compare the ground-truth answer against the predicted answer.
For example, suppose the ground-truth fact is:
\emph{“The man is wearing a red shirt.”}, while
the generated scene graph predicts: \emph{“A person wearing a dark red top.”}
We construct a verification question: \emph{“Does the generated description correctly state that the man is wearing a red shirt?”}
The LLM judge assigns a score from 0 (incorrect) to 5 (fully correct), allowing partial semantic credit. In practice, we curate specific judgment templates for each attribute or relation type and use Qwen3-235B as the judge.
We normalize the assigned score to $\tilde{g}_f$ and the overall Generation Score is: 
\begin{equation}
    G = \frac{1}{|F_g|} \sum_{f \in F_g} \tilde{g}_f,
\end{equation}

\noindent where $F_g$ is the set of generated atomic facts.

\subsubsection{Understanding Score}

For VQA answers $\hat{y}_k$, we use the same LLM-as-judge protocol to assign scores $u_k \in [0,5]$, which are then normalized.

The Understanding Score is:
\begin{equation}
    U = \frac{1}{K} \sum_{k=1}^{K} \tilde{u}_k.
\end{equation}

\subsubsection{Continuous Cross-Task Agreement}
Let $F$ denote the set of atomic facts shared between generation and
understanding. Since both tasks are evaluated on continuous $[0,1]$ scales,
we define a single consistency metric
\begin{equation}
    \text{CCTA} =
    \frac{\sum_{f \in F} w_f \left(1 - |\tilde{g}_f - \tilde{u}_f|\right)}
    {\sum_{f \in F} w_f},
    \label{eq:ccta}
\end{equation}
where $w_f = 1$ for simplicity. CCTA rewards symmetric responses and penalizes asymmetric ones. This design intentionally decouples consistency from accuracy, isolating pure representational symmetry. However, a model that is \emph{consistently wrong} is not desirable, even if it scores equally to the \emph{consistently correct} scenario. To penalize such coherent hallucination, we propose Accuracy-Weighted CCTA (AW-CCTA), which scales the consistency term by the mean joint accuracy:
\begin{equation}
    \text{AW-CCTA} =
    \frac{\displaystyle\sum_{f \in F} w_f \cdot
    \Bigl(1 - |\tilde{g}_f - \tilde{u}_f|\Bigr) \cdot
    \dfrac{\tilde{g}_f + \tilde{u}_f}{2}}
    {\displaystyle\sum_{f \in F} w_f}.
    \label{eq:aw_ccta}
\end{equation}

AW-CCTA assigns near-zero scores when both tasks are consistently wrong, and converges to CCTA when both tasks are consistently correct (i.e., $\tilde{g}_f, \tilde{u}_f \to 1$). The two metrics serve complementary diagnostic roles. CCTA measures whether a model behaves symmetrically across generation and understanding. AW-CCTA additionally requires that this symmetric behavior is grounded in factual accuracy, making it more meaningful for practical deployment. When the two metrics diverge significantly for a given model, it reveals that the model's apparent consistency stems from consistent hallucination and we empirically validate this diagnostic property through a human study in~\cref{sec:human}.

\section{Experiments}

\subsection{Experimental Setup}
We evaluate eight open-sourced uMMs with XTC-Bench, including BAGEL 7B~\cite{deng2025emergingpropertiesunifiedmultimodal}, BLIP3-o-8B~\cite{chen2025blip3ofamilyfullyopen}, Janus-Pro-7B~\cite{chen2025janusprounifiedmultimodalunderstanding}, MMaDA-8B~\cite{yang2025mmadamultimodallargediffusion}, OmniGen-2~\cite{wu2025omnigen2explorationadvancedmultimodal}, Show-o~\cite{xie2025showosingletransformerunify}, Show-o2-7B~\cite{xie2025showo2improvednativeunified}, and Tar-7B~\cite{han2025visiondialectunifyingvisual}. We chose those models to cover the taxonomy of uMM architectures proposed in recent work~\cite{zhao2026unifiedmultimodalunderstandinggeneration}. MMaDA leverages the diffusion paradigm for both visual and text modality. Tar, OmniGen2, BLIP3-o-8B, and Janus-Pro-7B adopt an autoregressive next-token prediction (NTP) backbone structure with sophisticated encoding and decoding strategies for both modalities. Show-o, Show-o2-7B, and BAGEL-7B predict text tokens with NTP and use diffusion for visual tokens. We also evaluate two commercial models \texttt{gemini-2.5-flash-image} and GPT (\texttt{gpt-image-1.5} $+$ \texttt{gpt-5}).\footnote{We only report the Generation and Understanding performance of GPT models as a strong reference for the two standalone tasks.}

\subsection{Evaluation Results}

We evaluate all models along three complementary dimensions: generation ($G$), understanding($U$), and cross-task consistency (CCTA and AW-CCTA).

\noindent\textbf{Generation and Understanding}~\cref{tab:G} and~\cref{tab:U} report per-task performance. Commercial models lead both dimensions: GPT-5/GPT-Image-1.5 achieves the highest overall generation score, while Gemini-2.5 Flash leads in understanding. Among open-source models, BAGEL-7B ranks first on both generation ($G{=}0.725$) and understanding ($U{=}0.724$). We also find that overall generation scores vary widely across models, yet matched-node attribute and relation scores converge substantially, with all open-source models scoring above $0.84$ on matched Rel.\ This reveals that overall generation gaps are driven primarily by models either missing generating the node object entirely or failing to ground the generated object into the ground truth rather than by semantic inaccuracies in the attributes or relations they do produce. In contrast, understanding scores exhibit no such convergence under matched evaluation. Matched relational understanding spans $0.330$ (MMaDA-8B) to $0.793$ (Gemini-2.5 Flash), indicating that relational comprehension represents a persistent semantic bottleneck.

\begin{table}[h]
\centering
\caption{Generation performance ($G$) evaluated on scene graph nodes. We report the performance for all nodes and matched nodes separately.}
\label{tab:G}
\scriptsize 
\setlength{\tabcolsep}{6pt} 

\begin{tabular}{l|cc|cc|cc}
\toprule
\textbf{Model} & \makecell[c]{\textbf{Overall}\\\textbf{Gen.}} & \makecell[c]{\textbf{Matched}\\\textbf{Nodes}} & \makecell[c]{\textbf{Attr.}} & \makecell[c]{\textbf{Rel.}} & \makecell[c]{\textbf{Matched}\\\textbf{Attr.}} & \makecell[c]{\textbf{Matched}\\\textbf{Rel.}} \\
\midrule
\makecell[c]{GPT-5 /\\GPT-Image-1.5}    & \boldfw{0.804} & \boldfw{89.0}\% & \boldfw{0.850} & \boldfw{0.757} & \boldfw{0.889} & \boldfw{0.917} \\
Gemini-2.5 Flash         & 0.795          & 87.6\%          & 0.842          & 0.749          & 0.888          & 0.906          \\
\midrule
BLIP3-o-8B               & 0.545          & 67.9\%          & 0.591          & 0.502          & 0.716          & 0.896          \\
Tar-7B                   & 0.518          & 62.5\%          & 0.564          & 0.479          & 0.743          & 0.902          \\
Janus-Pro-7B             & 0.499          & 62.2\%          & 0.557          & 0.455          & 0.799          & 0.908          \\
OmniGen-2                & 0.499          & \boldfw{80.1}\% & 0.593          & 0.411          & \boldfw{0.867} & 0.909          \\
Show-o                   & 0.060          & 14.3\%          & 0.083          & 0.041          & 0.703          & 0.893          \\
Show-o-2-7B              & 0.465          & 62.3\%          & 0.534          & 0.411          & 0.829          & 0.848          \\
BAGEL-7B                 & \boldfw{0.725} & 79.3\%          & \boldfw{0.782} & \boldfw{0.670} & 0.855          & \boldfw{0.912} \\
MMaDA-8B                 & 0.265          & 39.1\%          & 0.347          & 0.201          & 0.713          & 0.871          \\
\bottomrule
\end{tabular}
\end{table}

\begin{table}[h]
\centering
\caption{Understanding performance ($U$) measured via scene graph-derived VQA. We report the Overall Understanding score, followed by task-specific scores and accuracy restricted to matched nodes.}
\label{tab:U}
\scriptsize 
\setlength{\tabcolsep}{3pt} 
\begin{tabular}{l|c|ccc|cc}
\toprule
\textbf{Model} & \makecell[c]{\textbf{Overall}\\\textbf{Und.}} & \makecell[c]{\textbf{Obj.}\\\textbf{Retr.}} & \makecell[c]{\textbf{Attr.}\\\textbf{Query}} & \makecell[c]{\textbf{Rel.}\\\textbf{Query}} & \makecell[c]{\textbf{Matched}\\\textbf{Attr.}} & \makecell[c]{\textbf{Matched}\\\textbf{Rel.}} \\
\midrule
\makecell[c]{GPT-5 /\\GPT-Image-1.5}    & 0.684          & \boldfw{0.718} & 0.700          & 0.658          & 0.702          & 0.662 \\
Gemini-2.5 Flash         & \boldfw{0.740} & 0.686          & \boldfw{0.708} & \boldfw{0.789} & \boldfw{0.710} & \boldfw{0.793} \\
\midrule
BLIP3-o-8B               & 0.692          & 0.679          & 0.677          & 0.709          & 0.688          & 0.740          \\
Tar-7B                   & 0.580          & 0.557          & 0.591          & 0.580          & 0.612          & 0.615          \\
Janus-Pro-7B             & 0.681          & 0.616          & 0.665          & 0.719          & 0.682          & 0.745          \\
OmniGen-2                & 0.659          & 0.672          & 0.666          & 0.649          & 0.688          & 0.695          \\
Show-o                   & 0.503          & 0.415          & 0.579          & 0.467          & 0.632          & 0.582          \\
Show-o-2-7B              & 0.605          & 0.583          & 0.664          & 0.560          & 0.677          & 0.589          \\
BAGEL-7B                 & \boldfw{0.724} & \boldfw{0.674} & \boldfw{0.704} & \boldfw{0.759} & \boldfw{0.711} & \boldfw{0.767} \\
MMaDA-8B                 & 0.380          & 0.377          & 0.484          & 0.289          & 0.515          & 0.330          \\
\bottomrule
\end{tabular}
\end{table}

\begin{table}[t]
\centering
\caption{Continuous Cross-Task Agreement (CCTA) and Accuracy-Weighted CCTA (AW-CCTA) evaluated on all nodes.}
\label{tab:ccta_awccta_results}
\scriptsize
\setlength{\tabcolsep}{5pt}

\begin{tabular}{l|ccc|ccc}
\toprule
\textbf{Model} 
& \multicolumn{3}{c|}{\textbf{CCTA}} 
& \multicolumn{3}{c}{\textbf{AW-CCTA}} \\
& \textbf{Overall} 
& \textbf{Attributes} 
& \textbf{Relations} 
& \textbf{Overall} 
& \textbf{Attributes} 
& \textbf{Relations} \\
\midrule
Gemini-2.5 Flash         
    & \textbf{0.706} & 0.708 & \textbf{0.705} & \textbf{0.623} & 0.620 & \textbf{0.625} \\
\midrule
BLIP3-o-8B               
    & 0.598 & 0.625 & 0.573 & 0.423 & 0.448 & 0.400 \\
Tar-7B                   
    & 0.600 & 0.630 & 0.574 & 0.358 & 0.394 & 0.327 \\
Janus-Pro-7B             
    & 0.573 & 0.610 & 0.544 & 0.390 & 0.415 & 0.370 \\
OmniGen-2                
    & 0.585 & 0.615 & 0.557 & 0.377 & 0.439 & 0.320 \\
Show-o                   
    & 0.509 & 0.461 & 0.548 & 0.054 & 0.068 & 0.042 \\
Show-o-2-7B              
    & 0.582 & 0.592 & 0.574 & 0.333 & 0.395 & 0.285 \\
BAGEL-7B                 
    & \boldfw{0.675} & \boldfw{0.696} & 0.654 & \boldfw{0.569} & \boldfw{0.588} & \boldfw{0.550} \\
MMaDA-8B                 
    & 0.630 & 0.590 & \boldfw{0.662} & 0.144 & 0.215 & 0.088 \\
\bottomrule
\end{tabular}
\end{table}

\begin{wrapfigure}{r}{0.48\columnwidth}
    \centering
    \vspace{-10pt} 
    \includegraphics[width=\linewidth]{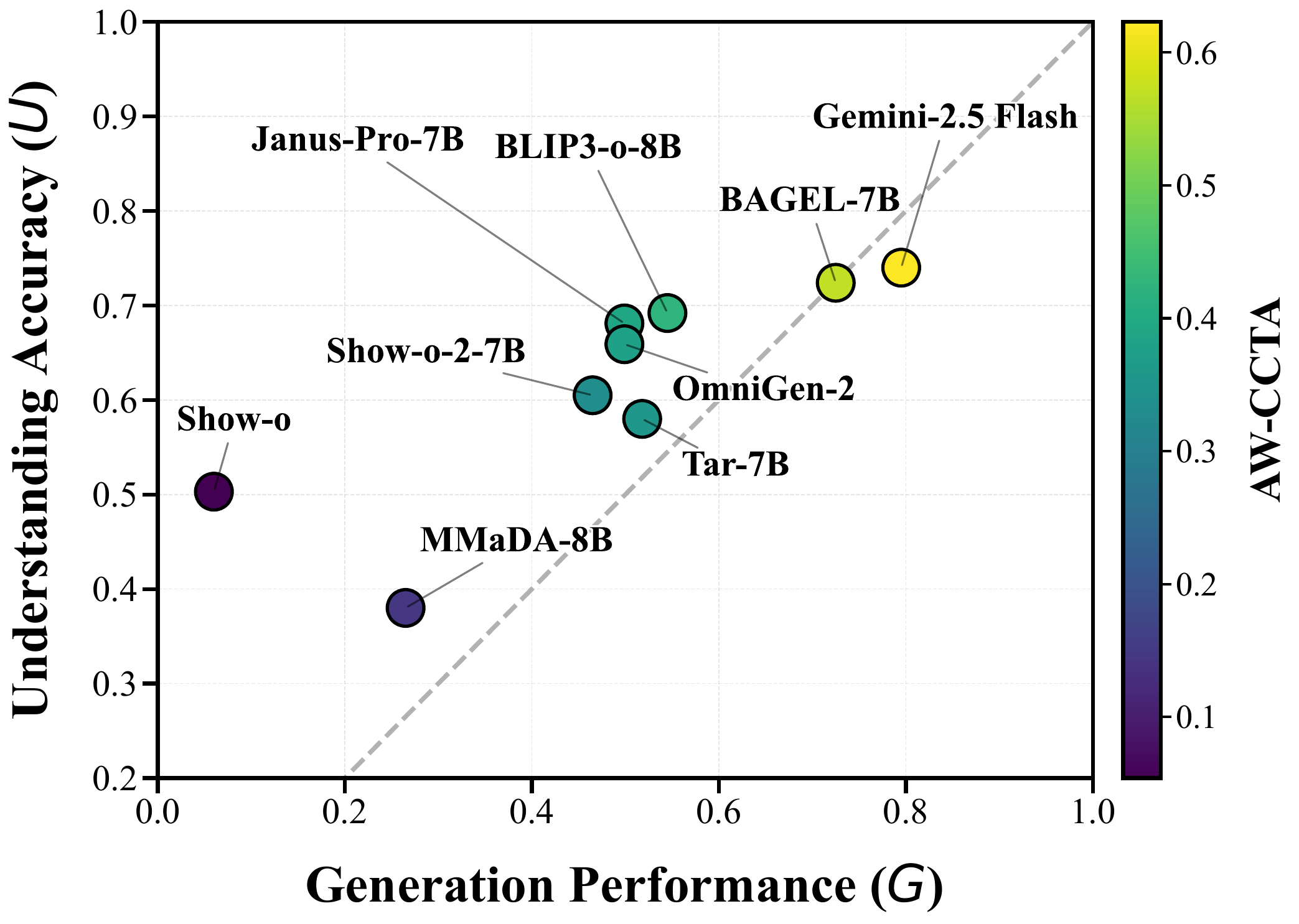}
    \caption{Task performance ($G, U$) versus semantic consistency (AW-CCTA).}
    \label{fig:correlation}
    \vspace{-10pt} 
\end{wrapfigure}

\cref{tab:ccta_awccta_results} reports CCTA and AW-CCTA results. First, cross-task inconsistency is observed universally: the strongest CCTA reaches only $0.706$, and under AW-CCTA the best model achieves $0.623$, demonstrating that semantic misalignment is a structural property not resolved by scale or architectural sophistication. Second, the CCTA--AW-CCTA gap exposes a failure mode invisible to per-task metrics. MMaDA-8B ranks third on raw CCTA ($0.630$) despite near-bottom $G$ and $U$, yet its AW-CCTA collapses to $0.144$—the diagnostic signature of consistent hallucination, where both tasks produce correlated incorrect responses that yield spuriously high raw agreement. Neither $G$ nor $U$ alone can detect this; however, the CCTA--AW-CCTA divergence can directly quantify it.

\cref{fig:correlation} plots $G$ against $U$ with color encoding AW-CCTA. Nearly all models lie above the parity line, indicating that understanding systematically outpaces generation, suggesting that current training objectives favor perceptual grounding over generative faithfulness. Models closest to the parity line (BAGEL-7B, Gemini-2.5 Flash) achieve the highest AW-CCTA, suggesting that balanced task performance is a necessary, if not sufficient, condition for cross-task alignment. We also find that neither $G$ nor $U$ alone reliably predicts AW-CCTA, confirming that consistency constitutes an independent evaluation dimension.

To further examine whether this pattern is architecturally structured, \cref{tab:arch_family} aggregates results by family. Pure Diffusion (MMaDA) achieves the smallest G--U gap and high raw CCTA, yet collapses under AW-CCTA, revealing symmetric failure rather than genuine alignment. Pure AR achieves the highest mean AW-CCTA ($0.387$), suggesting that a unified next-token prediction objective naturally couples the two tasks. AR+Diffusion models exhibit the largest mean G--U gap ($-0.194$) and a 
$10{\times}$ within-group AW-CCTA spread ($0.054$--$0.569$), exposing a strong sensitivity to the calibration between the NTP and diffusion components. \textbf{Taken together, these results indicate that cross-task consistency is governed not by architectural unification alone, but by whether the learning objectives across modalities are sufficiently coupled. This is a property that is invisible to per-task metrics and is only surfaced by explicit consistency measurement.}

\begin{table}[h]
\centering
\caption{Architecture family averages.}
\label{tab:arch_family}
\scriptsize
\resizebox{\columnwidth}{!}{
\begin{tabular}{l|ccccc}
\toprule
\textbf{Family} & \textbf{Mean G} & \textbf{Mean U} & \textbf{G--U Gap} 
& \textbf{Mean CCTA} & \textbf{Mean AW-CCTA} \\
\midrule
Pure Diffusion (MMaDA)               & 0.265 & 0.380 & $-$0.115 & 0.630 & 0.144 \\
AR+Diffusion (Show-o, Show-o2, BAGEL)& 0.417 & 0.611 & $-$0.194 & 0.589 & 0.319 \\
Pure AR (OmniGen-2, BLIP3-o, Janus-Pro, Tar) & 0.515 & 0.653 & $-$0.138 & 0.589 & 0.387 \\
\bottomrule
\end{tabular}
}
\end{table}

\subsection{Reproducibility}

To assess reproducibility under limited resources, we re-ran the full protocol with smaller Qwen3-VL variants for all three judge-dependent components (attribute generation, relation validation, and VQA scoring). We report results for open-source generation models only and exclude proprietary baselines from this analysis.

Across model scales, the main conclusions remain stable. BAGEL-7B remains the strongest open-source model under AW-CCTA, while MMaDA-8B continues to exhibit the same ``consistent but wrong'' pattern (relatively high raw CCTA with low AW-CCTA). Quantitatively, averaging over open-source models, replacing the 235B judge with the 32B variant changes $G$ from 0.447 to 0.436, $U$ from 0.603 to 0.542, CCTA from 0.594 to 0.623, and AW-CCTA from 0.331 to 0.301. For 7B, the corresponding values are $G=0.389$, $U=0.534$, CCTA$=0.603$, and AW-CCTA$=0.273$.

These results indicate that the benchmark remains reproducible with smaller judge models. Absolute scores are lower, especially for AW-CCTA at 7B, but diagnostic trends and comparative ordering among open models are largely preserved. This supports the idea of using 32B/7B as a resource-friendly alternative for reproduction while retaining 235B as the highest-fidelity reference.

\begin{table}[h]
\centering
\caption{Open-source aggregate means across protocol scales.}
\label{tab:scale_agg}
\small
\begin{tabular}{l|ccccc}
\toprule
\textbf{Scale} & \makecell[c]{\textbf{Matched}\\\textbf{Nodes (\%)}} & \textbf{Mean G} & \textbf{Mean U} & \textbf{CCTA} & \textbf{AW-CCTA} \\
\midrule
Qwen3-VL-235B & 58.5\% & 0.447 & 0.603 & 0.594 & 0.331 \\
Qwen3-VL-32B  & 58.5\% & 0.436 & 0.542 & 0.623 & 0.301 \\
Qwen3-VL-7B   & 58.5\% & 0.389 & 0.534 & 0.603 & 0.273 \\
\bottomrule
\end{tabular}
\end{table}

The complete per-model results for both protocol scales are reported in the appendix tables.

\subsection{Human Study}
\label{sec:human}

To validate the reliability of XTC-Bench’s automated dataset construction and evaluation framework, we conducted human studies on four components: (i) reference scene graph quality, (ii) prompt fidelity, (iii) LLM-judge reliability, and (iv) CCTA human alignment. All evaluations are performed at the level of atomic facts (objects, attributes, and relationships). 
Five master’s-level computer science students independently evaluated balanced subsets of 200 samples for each experiment. Aggregated results are reported in~\cref{tab:human_eval_split} and~\cref{tab:decoupled_validation}.

\begin{table}[h]
    \centering
    \caption{Human evaluation results split by category. Scene graph and prompt accuracy are computed as $\bar{s}/5 \times 100$\,\% (0--5 scale); LLM judge agreement is computed using Pearson correlation scores. Values are means over annotators.}
    \label{tab:human_eval_split}
    \begin{tabular}{lcccc}
        \toprule
        \textbf{Evaluation} & \textbf{Objects} & \textbf{Attributes} & \textbf{Relationships} & \textbf{Overall} \\
        \midrule
        Scene Graph Eval (Acc. \%) & \hphantom{0}94.44 & 92.48 & 95.17 & 94.02 \\
        Prompt Eval (Acc. \%) & 100.00 & 99.40 & 92.08 & 97.15 \\
       LLM Judge Eval ($\rho$) & 0.928 & 0.800 & 0.850 & 0.892 \\
        \bottomrule
    \end{tabular}
\end{table}

\subsubsection{Scene Graph Quality}
Here, the annotators were requested to assess the correctness of extracted objects, attributes, and relations using segmentation mask overlays and score each fact on a 0–5 scale. The pipeline achieves an overall accuracy of 94.02\%, with strong performance across all aspects, confirming the reliability of our automated scene graph extraction.

\subsubsection{Prompt Fidelity}
In this experiment, annotators evaluated whether generated prompts faithfully preserve scene graph facts. The overall accuracy reaches 97.15\%, with near-perfect coverage for objects (100.00\%) and attributes (99.40\%), demonstrating that prompt linearization and refinement maintain structural consistency with the underlying scene graph.

\subsubsection{LLM Judge Reliability}
To assess automated scoring quality, annotators evaluated model outputs on a 0–5 scale to measure alignment with the LLM judge. The overall correlation reaches $\rho=0.892$, with strong correlations across all categories, demonstrating that the LLM reliably preserves relative quality rankings consistent with human judgment.

\subsubsection{CCTA Alignment}

To evaluate the capability of the metrics to determine how well they measure consistency and correctness, we ask annotators to assign two independent Likert scores per fact triplet, both judged with access to the reference fact: consistency ($h_{\text{con}}$), assessing whether the outputs are consistently correct or consistently hallucinated, and semantic reliability ($h_{\text{rel}}$), assessing the correctness of both understanding and generation. This two-question design ensures methodologically independent validation of CCTA and AW-CCTA, respectively.

\cref{tab:decoupled_validation} reports Spearman correlations against human axes. Raw CCTA achieves strong consistency correlation ($\rho_{\text{con}}{=}0.79$) but exhibits a severe divergence on reliability ($\rho_{\text{rel}}{=}0.39$, $\Delta{=}0.39$). This empirical gap confirms that pure consistency is an insufficient proxy for model utility, as it rewards the consistent hallucination (\texttt{AH}) failure mode equally to the consistent correctness (\texttt{AC}) success mode. This trend holds across fact types, with relations showing the weakest correlation for raw CCTA ($\rho_{\text{rel}}{=}0.26$). In contrast, AW-CCTA reduces this gap ($\Delta{=}0.30$) while exhibiting a respectable consistency correlation ($\rho_{\text{con}}{=}0.57$) and showing high reliability ($\rho_{\text{rel}}=0.86$) breakdowns across attributes ($0.92$) and relations ($0.86$). The two metrics are thus complementary: CCTA diagnoses representational symmetry independently of correctness, while AW-CCTA reflects deployment-relevant alignment grounded in factual accuracy.

\begin{table}[t]
    \centering
    \small
    \caption{Spearman correlation ($\rho$) against human judgment. We compare \textbf{Consistency} vs. \textbf{Reliability}. The last three columns are fixed to equal width for clear comparison.}
    \label{tab:decoupled_validation}
    
    \resizebox{0.8\linewidth}{!}{
        \begin{tabular}{l c c r >{\centering\arraybackslash}p{1.7cm} >{\centering\arraybackslash}p{1.7cm}}
            \toprule
            & \textbf{Consistency} & \textbf{Reliability} & \multicolumn{1}{c}{\textbf{Gap}} & \multicolumn{2}{c}{\textbf{Reliability Breakdown}} \\
            \cmidrule(lr){5-6}
            \textbf{Metric} & ($\rho_{\text{con}}$) & ($\rho_{\text{rel}}$) & \multicolumn{1}{c}{($\Delta$)} & \textbf{Attr} & \textbf{Rel}\\
            \midrule
            \textbf{CCTA (Raw)}  & \textbf{0.79} & 0.39 & $-0.39$ & 0.56 & 0.26 \\
            \textbf{AW-CCTA}     & 0.57 & \textbf{0.86} & $\hphantom{-}0.30$ & \textbf{0.92} & \textbf{0.86} \\
            \bottomrule
        \end{tabular}
    }
\end{table}

\section{Limitations and Future Work}

While human evaluation validates the reliability of XTC-Bench, our framework relies on automated teacher models (e.g., Fair-PSG and Qwen3-VL) for scene graph extraction, attribute prediction, and final metric computation. This automated pipeline may inherit the specific biases of these teacher models or systematically fail when processing highly abstract or non-naturalistic visual concepts. Future iterations of the benchmark could incorporate more diverse judge ensembles or integrate human-in-the-loop scoring for highly ambiguous scenes to continuously calibrate the automated metrics. The current attribute schema could also be refined to better capture object-specific properties that are important for cross-task consistency, such as age, gender presentation, or other semantically salient fine-grained characteristics.

Furthermore, XTC-Bench operates strictly as a black-box evaluation protocol. It quantifies the behavioral misalignment between generation and understanding but does not isolate the internal mechanisms driving this discrepancy. Future work should integrate white-box probing techniques to identify precisely where within the unified architecture---such as specific transformer layers or modality routers---the semantic representations of perception and synthesis begin to diverge.

In addition, the current formulation of AW-CCTA assigns a uniform weight to every atomic fact, implicitly assuming that all facts contribute equally to cross-task consistency. This simplification may overlook that some facts are more visually apparent than others, for example because they correspond to large objects, persistent scene elements, or attributes with strong perceptual significance. A promising direction for future work is therefore to investigate non-uniform weighting schemes that better reflect the relative importance of different facts for the final image.

The current framework also represents objects primarily through a loose relational world model, without anchoring them to explicit spatial coordinates or distance-based structure in the image plane. Incorporating stronger spatial grounding could enable comparisons that account not only for semantic equivalence but also for structural correspondence between the reference and generated images.

Additionally, the current framework is constrained to static images and text. As unified multimodal models expand to encompass native video, audio, and spatial computing, future research must extend this structured evaluation framework, potentially utilizing dynamic spatio-temporal scene graphs to measure cross-task consistency in any-to-any multimodal systems.

Finally, while this study systematically diagnoses representational inconsistency, it does not propose an architectural or training-based solution. We anticipate that our findings will motivate future research into explicit consistency constraints during pre-training, novel bidirectional loss functions, or inference-time self-verification algorithms designed to actively enforce semantic alignment across tasks.

\section{Conclusion}
We present \textbf{XTC-Bench}, a scene-graph-grounded benchmark for evaluating cross-task visual semantic consistency in unified Multimodal Models. By grounding both generation prompts and understanding queries in a shared structured scene graph, XTC-Bench enables fine-grained, fact-level comparison across objects, attributes, and relations. We further propose CCTA and AW-CCTA to disentangle representational symmetry from factual accuracy, exposing coherent hallucination as a failure mode that per-task 
metrics cannot detect. Experiments reveal that even state-of-the-art models exhibit systematic misalignment, particularly at the attribute and relational levels, and that cross-task consistency is governed not by architectural unification but by the degree of coupling between modality-specific learning objectives. We acknowledge that automated scene-graph extraction and 
LLM-based evaluation introduce minor noise, though our human study confirms this is negligible. We envision XTC-Bench as a principled foundation for future research toward unified models that maintain coherent semantic representations across both perception and synthesis.

\section*{Acknowledgements}
The project on which this publication is based was funded by the Federal Ministry of Research, Technology and Space under the funding code “KI-Servicezentrum Berlin-Brandenburg” 16IS22092. Responsibility for the content of this publication remains with the author.

\bibliographystyle{splncs04}
\bibliography{main}

\section*{Appendix}

\subsection{Detailed Dataset Implementation}

\subsubsection{Selection of Panoptic Segmentation Model}

As described in Section 3.2.1 of the main paper, our approach builds upon the two-stage Fair-PSGG framework for panoptic scene graph generation. In the first stage, the input image is decomposed into a set of object and stuff regions with associated semantic labels. These panoptic predictions form the structural basis for the subsequent relationship reasoning stage. As a result, the quality of the panoptic segmentation directly affects the overall scene graph generation performance. Prior work \cite{lorenz2024fairpsgg} shows that improvements in panoptic segmentation accuracy translate into consistent gains in downstream scene graph metrics.

To ensure a strong and reliable foundation for our framework, we evaluate several recent state-of-the-art panoptic segmentation models. \Cref{tab:segmentation_comparison} reports their performance on the COCO val2017 split using the standard Panoptic Quality (PQ) metric, together with its decomposition into \textit{Things} (PQ$_{Th}$) and \textit{Stuff} (PQ$_{St}$) categories. These metrics capture segmentation and recognition quality jointly and are widely used to assess panoptic segmentation performance.

Among the compared approaches, \textbf{kMaX-DeepLab} achieves a highly competitive overall PQ of 58.1 while providing the strongest performance on \textit{Stuff} categories with a PQ$_{St}$ of 48.8. Accurate modeling of stuff regions is particularly important for scene graph generation, since contextual relations frequently involve background elements (e.g., \textit{road}, \textit{sky}, or \textit{grass}). At the same time, the model maintains strong performance on \textit{Things} categories (PQ$_{Th}$ = 64.3), ensuring reliable instance-level segmentation (\cref{tab:segmentation_comparison}).

Given its balanced performance across both object and background classes, we select \textbf{kMaX-DeepLab} as the panoptic backbone for all experiments in this work.

\begin{table}[h]
\centering
\caption{Comparison of state-of-the-art panoptic segmentation models on the COCO val2017 unlabeled dataset. We report Overall Panoptic Quality (PQ), as well as PQ for Things (PQ$_{Th}$) and Stuff (PQ$_{St}$).}
\label{tab:segmentation_comparison}

\begin{tabular}{l|ccc}
\toprule
\textbf{Model} & \textbf{PQ overall} & \textbf{PQ$_{Th}$} & \textbf{PQ$_{St}$} \\
\midrule
MaX-DeepLab~\cite{wang2021maxdeeplabendtoendpanopticsegmentation} & 51.1 & 57.0 & 42.2 \\
MaskFormer~\cite{cheng2021perpixelclassificationneedsemantic} & 52.7 & 58.5 & 44.0 \\
K-Net~\cite{zhang2021knetunifiedimagesegmentation} & 54.6 & 60.2 & 46.0 \\
CMT-DeepLab~\cite{yu2022cmtdeeplabclusteringmasktransformers} & 55.3 & 61.0 & 46.6 \\
Panoptic SegFormer~\cite{li2022panopticsegformerdelvingdeeper} & 55.8 & 61.7 & 46.9 \\
Mask2Former~\cite{cheng2022maskedattentionmasktransformeruniversal} & 57.8 & 64.2 & 48.1 \\
OneFormer~\cite{jain2022oneformertransformerruleuniversal} & 57.9 & \boldfw{64.4} & 48.0 \\
\textbf{kMax-DeepLab}~\cite{yu2023kmaxdeeplabkmeansmasktransformer} & \boldfw{58.1} & 64.3 & \boldfw{48.8} \\
\bottomrule
\end{tabular}
\end{table}

As described in Section 3.2.1 of the main paper, our scene graph generation framework operates on a predefined set of predicates that capture various interactions between entities. To facilitate structured evaluation, we group these predicates into a four-part taxonomy based on their relational characteristics:

\begin{itemize}
\item \textbf{Spatial:} Captures positional and support relations such as \textit{above, in, on, next to, under,} and \textit{behind}.
\item \textbf{Posture:} Describes the physical orientation or stance of a subject, including \textit{standing on, sitting on, lying on,} and \textit{leaning against}.
\item \textbf{Locomotion:} Involves movement-based interactions such as \textit{walking on, running on, riding, flying over,} and \textit{moving towards}.
\item \textbf{Social:} Represents high-level human or animal interactions, including \textit{looking at, holding, talking to, playing with,} and \textit{following}.
\end{itemize}

This categorization reflects the primary interaction patterns observed in natural scenes and allows for a more interpretable comparison of model performance across physical, directional, and behavioral dimensions.

\subsubsection{Performance of Panoptic Scene Graph Generation}

To evaluate the effectiveness of our pipeline, we trained the Decoupled SceneFormer (DSFormer) architecture following the two-stage Fair-PSGG setup. Since the original model weights were not publicly released, we reproduced the training procedure using the Microsoft COCO dataset. This ensures that our evaluation remains consistent with the experimental protocol described by the original work~\cite{lorenz2024fairpsgg}. For benchmarking, we evaluated the trained model on the COCO val2017 split, which provides a standardized evaluation protocol for panoptic scene graph generation. The results, reported per predicate in~\cref{tab:predicate_stats_refined}, demonstrate performance trends that closely match those reported in the original Fair-PSGG paper. In particular, the Predicate Classification (PredCls) setting achieves a mean accuracy of 0.77 across all predicates, confirming the successful reproduction of the training procedure. Minor deviations from the original paper can be attributed to differences in training initialization and stochastic optimization.

\begin{table}[!ht]
\centering
\caption{Detailed predicate performance metrics. We report accuracy (Acc), balanced accuracy (Bal\_Acc), F1-score, and Mean Recall ($mR@K$) at various thresholds.}
\label{tab:predicate_stats_refined}
\small
\begin{adjustbox}{totalheight=\textheight-6\baselineskip}
\setlength{\tabcolsep}{4pt} 
\begin{tabularx}{1\textwidth}{l|ccc|cccc}
\toprule
\textbf{Predicate} & \textbf{Acc} & \textbf{Bal\_Acc} & \textbf{F1} & \textbf{mR@10} & \textbf{mR@20} & \textbf{mR@50} & \textbf{mR@-1} \\
\midrule
mean & 0.77 & 0.77 & 0.18 & 0.27 & 0.34 & 0.39 & 0.21 \\
\hline
NONE & 0.49 & 0.49 & 0.03 & --- & --- & --- & --- \\
\hline
over & 0.88 & 0.88 & 0.46 & 0.58 & 0.67 & 0.73 & 0.49 \\
in front of & 0.70 & 0.70 & 0.30 & 0.15 & 0.26 & 0.40 & 0.07 \\
beside & 0.55 & 0.55 & 0.13 & 0.07 & 0.14 & 0.22 & 0.03 \\
on & 0.67 & 0.67 & 0.40 & 0.31 & 0.34 & 0.37 & 0.27 \\
in & 0.80 & 0.80 & 0.30 & 0.23 & 0.28 & 0.32 & 0.18 \\
attached to & 0.59 & 0.59 & 0.18 & 0.16 & 0.23 & 0.27 & 0.09 \\
hanging from & 0.91 & 0.91 & 0.26 & 0.48 & 0.57 & 0.68 & 0.39 \\
on back of & 0.63 & 0.63 & 0.05 & 0.05 & 0.13 & 0.15 & 0.03 \\
falling off & 0.67 & 0.67 & 0.07 & 0.33 & 0.33 & 0.33 & 0.33 \\
going down & 0.80 & 0.79 & 0.04 & 0.33 & 0.47 & 0.47 & 0.13 \\
painted on & 0.69 & 0.69 & 0.02 & 0.22 & 0.50 & 0.50 & 0.05 \\
walking on & 0.88 & 0.88 & 0.18 & 0.33 & 0.44 & 0.49 & 0.27 \\
running on & 0.93 & 0.93 & 0.15 & 0.47 & 0.49 & 0.49 & 0.47 \\
crossing & 0.63 & 0.63 & 0.04 & 0.10 & 0.16 & 0.29 & 0.07 \\
standing on & 0.87 & 0.87 & 0.33 & 0.30 & 0.33 & 0.34 & 0.25 \\
lying on & 0.88 & 0.88 & 0.21 & 0.33 & 0.35 & 0.37 & 0.32 \\
sitting on & 0.83 & 0.83 & 0.34 & 0.34 & 0.40 & 0.47 & 0.29 \\
flying over & 0.86 & 0.86 & 0.17 & 0.38 & 0.52 & 0.68 & 0.17 \\
jumping over & 0.73 & 0.72 & 0.10 & 0.25 & 0.25 & 0.25 & 0.19 \\
jumping from & 0.74 & 0.75 & 0.16 & 0.15 & 0.15 & 0.15 & 0.15 \\
wearing & 0.88 & 0.89 & 0.43 & 0.50 & 0.61 & 0.63 & 0.40 \\
holding & 0.90 & 0.90 & 0.56 & 0.43 & 0.50 & 0.54 & 0.37 \\
carrying & 0.94 & 0.94 & 0.56 & 0.58 & 0.70 & 0.79 & 0.52 \\
looking at & 0.70 & 0.70 & 0.21 & 0.04 & 0.08 & 0.13 & 0.03 \\
guiding & 0.66 & 0.66 & 0.02 & 0.08 & 0.15 & 0.23 & 0.08 \\
kissing & --- & --- & --- & 0.50 & 1.00 & 1.00 & 0.00 \\
eating & 0.93 & 0.93 & 0.27 & 0.44 & 0.46 & 0.46 & 0.36 \\
drinking & 0.75 & 0.75 & 0.05 & 0.43 & 0.57 & 0.71 & 0.29 \\
feeding & 1.00 & 1.00 & 0.07 & 0.22 & 0.33 & 0.37 & 0.11 \\
biting & 0.89 & 0.88 & 0.10 & 0.50 & 0.50 & 0.50 & 0.46 \\
catching & 0.75 & 0.75 & 0.07 & 0.07 & 0.07 & 0.07 & 0.00 \\
picking & 0.50 & 0.50 & 0.00 & 0.00 & 0.00 & 0.00 & 0.00 \\
playing with & 0.75 & 0.75 & 0.02 & 0.03 & 0.22 & 0.08 & 0.03 \\
chasing & 0.66 & 0.66 & 0.02 & 0.00 & 0.00 & 0.11 & 0.00 \\
climbing & 0.75 & 0.75 & 0.04 & 0.25 & 0.25 & 0.25 & 0.25 \\
cleaning & 0.60 & 0.60 & 0.07 & 0.00 & 0.00 & 0.00 & 0.00 \\
playing & 0.95 & 0.95 & 0.38 & 0.55 & 0.57 & 0.59 & 0.51 \\
touching & 0.69 & 0.68 & 0.07 & 0.12 & 0.14 & 0.20 & 0.10 \\
pushing & 0.62 & 0.62 & 0.01 & 0.17 & 0.17 & 0.17 & 0.08 \\
pulling & 0.75 & 0.75 & 0.05 & 0.14 & 0.14 & 0.20 & 0.09 \\
opening & 0.50 & 0.50 & 0.00 & 0.00 & 0.00 & 0.00 & 0.00 \\
cooking & 1.00 & 1.00 & 0.30 & 0.60 & 0.60 & 0.80 & 0.60 \\
talking to & 0.91 & 0.91 & 0.06 & 0.04 & 0.17 & 0.38 & 0.02 \\
throwing & 0.93 & 0.93 & 0.16 & 0.00 & 0.00 & 0.00 & 0.00 \\
slicing & 0.99 & 1.00 & 0.08 & 0.33 & 0.42 & 0.50 & 0.13 \\
driving & 0.97 & 0.97 & 0.18 & 0.37 & 0.44 & 0.53 & 0.26 \\
riding & 0.82 & 0.82 & 0.51 & 0.30 & 0.34 & 0.37 & 0.26 \\
parked on & 0.96 & 0.96 & 0.34 & 0.46 & 0.54 & 0.59 & 0.38 \\
driving on & 0.96 & 0.96 & 0.39 & 0.61 & 0.67 & 0.68 & 0.54 \\
about to hit & 0.94 & 0.94 & 0.33 & 0.63 & 0.74 & 0.89 & 0.52 \\
kicking & 0.88 & 0.87 & 0.06 & 0.50 & 0.50 & 0.50 & 0.50 \\
swinging & 0.98 & 0.98 & 0.41 & 0.32 & 0.34 & 0.37 & 0.26 \\
entering & 0.66 & 0.66 & 0.01 & 0.03 & 0.09 & 0.26 & 0.03 \\
exiting & 0.75 & 0.75 & 0.02 & 0.20 & 0.20 & 0.40 & 0.20 \\
enclosing & 0.70 & 0.70 & 0.16 & 0.17 & 0.29 & 0.40 & 0.05 \\
leaning on & 0.74 & 0.74 & 0.08 & 0.09 & 0.14 & 0.18 & 0.07 \\
\bottomrule
\end{tabularx}
\end{adjustbox}
\end{table}

\subsubsection{Thresholding and Filtering Logic}

As described in Section 3.2.2 of the main paper, we apply confidence-based filtering to reduce noisy relation predictions produced by the scene graph generation. This step prioritizes precision over recall, ensuring that only high-confidence relationships are passed to the subsequent refinement stage. Specifically, we apply two thresholds. First, predictions classified as \textit{No Relation} are filtered using a confidence threshold of $s_{ij}^{NR} = 0.5$. If the model assigns a confidence score above this threshold to the \textit{No Relation} class for a subject--object pair $(i,j)$, the pair is discarded from further processing. This prevents spurious object pairs from entering the relation refinement stage. Second, candidate predicates must satisfy a minimum confidence threshold of $s_{ij}^{p} \geq 0.4$. Only predicates whose predicted probability exceeds this threshold are retained as potential relations. This filtering stage significantly reduces the number of low-confidence predictions while preserving the majority of meaningful relationships. Together, these thresholds create a balanced trade-off between computational efficiency and relational accuracy, ensuring that the downstream LLM-based verification stage operates on a curated set of high-quality candidate relations.

\subsection{LLM Judge and Prompting Protocols}

\subsubsection{Relation Refinement}

To further improve relational accuracy, we employ a vision-language model as a verification module that evaluates candidate relations produced by the scene graph generator. Specifically, we use Qwen3-VL-235B to perform pairwise relation validation between subject and object instances. For each candidate relation, the model receives the original image together with two highlighted bounding boxes: the \textit{subject} (marked in red) and the \textit{object} (marked in blue). The model is then asked to verify whether the predicted relationship is visually valid.

The following prompt is used for relation verification:

\begin{quote}
\textit{Prompt:} ``Verify the relationship between the object in the RED bounding box (Subject) and the BLUE bounding box (Object). Return JSON with key `answer` set to `Yes` or `No` only.''
\end{quote}

The constrained JSON output format ensures deterministic parsing of the model's response and minimizes ambiguity during automated evaluation. Relations confirmed with a \texttt{Yes} response are retained in the final scene graph, while those receiving a \texttt{No} response are discarded.

\subsubsection{Attribute Generation Meta-classes}

As described in the main paper, we introduce \textit{attribute meta-classes} to group COCO object categories that share semantically similar attribute types. This strategy reduces the number of prompts required for attribute generation while ensuring that queried attributes remain contextually appropriate for the object category. Each meta-class defines a set of attribute keys that are relevant to the associated objects. For example, electronic devices such as \textit{cell phone}, \textit{laptop}, and \textit{television} share the attribute \textit{screen state} (e.g., \textit{screen on/off}). In contrast, household appliances such as \textit{refrigerator} and \textit{oven} are evaluated using attributes such as \textit{door open/closed}. Similarly, clothing items share attributes related to \textit{primary color}, while food-related objects may include attributes describing \textit{topping type}.

By organizing object categories into meta-classes, we ensure that attribute queries remain semantically meaningful while avoiding unnecessary prompt duplication. The complete mapping of meta-classes to their corresponding COCO object categories and associated attribute keys is provided in Table~\ref{tab:attribute_metaclasses}.

\begin{table}[!ht]
\centering
\caption{Mapping of super categories to COCO classes / corresponding attribute keys.}
\label{tab:attribute_metaclasses}
\centering
\small
\resizebox{1.0\textwidth}{!}{
\begin{tabularx}{1.4\textwidth}{l>{\raggedright\arraybackslash}X>{\raggedright\arraybackslash}X}
\toprule
\textbf{Supercategory} & \textbf{COCO Categories} & \textbf{Attribute Keys} \\ \midrule
person & person & upper clothing type/color, lower clothing type/color, held object type, headwear/eyewear \\ \midrule
vehicle & airplane, bicycle, boat, bus, car, motorcycle, train, truck & primary color, viewpoint angle, text/number visible \\ \midrule
traffic control & traffic light, stop sign, parking meter & light color state, text on sign, mounted position \\ \midrule
street furniture & bench, fire hydrant & primary color, material type \\ \midrule
animal & bear, bird, cat, cow, dog, elephant, giraffe, horse, sheep, zebra & primary color, pattern type, body position \\ \midrule
bags accessories & backpack, handbag, suitcase, umbrella & primary color, pattern type, open/closed state \\ \midrule
wearables & tie & primary color, pattern type \\ \midrule
sports equipment & baseball bat, baseball glove, frisbee, kite, skateboard, skis, snowboard, sports ball, surfboard, tennis racket & primary color, brand/text visible \\ \midrule
containers & bottle, bowl, cup, wine glass & primary color, material type, content visible \\ \midrule
utensils & fork, knife, spoon & primary color, material type \\ \midrule
prepared food & cake, donut, hot dog, pizza, sandwich, food-other-merged & primary color, topping type \\ \midrule
raw produce & apple, banana, broccoli, carrot, fruit, orange & primary color \\ \midrule
furniture & bed, chair, couch, dining table & primary color, material type \\ \midrule
potted veg. & potted plant, flower & primary color, container type \\ \midrule
landscape veg. & grass-merged, tree-merged & primary color \\ \midrule
screen devices & cell phone, laptop, tv & primary color, brand/text visible, screen on/off \\ \midrule
input devices & keyboard, mouse, remote & primary color, brand/text visible \\ \midrule
door appliances & microwave, oven, refrigerator, toaster & primary color, material type, door open/closed \\ \midrule
bathroom & sink, toilet & primary color, material type \\ \midrule
indoor objects & book, clock, hair drier, scissors, teddy bear, toothbrush, vase & primary color, text visible, content visible \\ \midrule
textiles & banner, blanket, curtain, pillow, towel, rug-merged & primary color, pattern type, text visible \\ \midrule
surfaces & road, pavement-merged, sand, gravel, snow, dirt-merged & surface material, marking type, wet/dry state \\ \midrule
buildings & building-other-merged, house, tent & primary color, material type, window count \\ \midrule
infrastructure & bridge, fence-merged, stairs, roof, platform, railroad & primary color, material type \\ \midrule
openable & door-stuff, window-blind, window-other & material type, primary color, open/closed state \\ \midrule
room surfaces & ceiling-merged, floor-other-merged, floor-wood, wall-brick, wall-other-merged, wall-stone, wall-tile, wall-wood & material type, primary color \\ \midrule
storage & cabinet-merged, shelf & primary color, material type, drawer count \\ \midrule
work surfaces & table-merged, counter & primary color, material type \\ \midrule
outdoor elem. & cardboard, light, mirror-stuff, net, paper-merged, playingfield, rock-merged & primary color, material type, text visible \\ \midrule
natural env. & mountain-merged, river, sea, sky-other-merged, water-other & primary color, weather type, wave/cloud visible \\ \bottomrule
\end{tabularx}
}
\end{table}

\subsubsection{Visual Chain-of-Thought (CoT) Prompt Template}

For attribute generation, we employ structured prompting with Qwen3-VL-235B using a visual Chain-of-Thought (CoT) formulation. The model receives two complementary visual inputs: (1) a \textit{Detail Crop} focusing on the target object instance and (2) a \textit{Semantic Spotlight} image highlighting the object within the broader scene context. This dual-view setup allows the model to reason about both local appearance and surrounding contextual cues.

The model is instructed to analyze a predefined set of attribute keys associated with the object's meta-class. To encourage structured reasoning and improve attribute reliability, we request intermediate reasoning steps prior to producing the final structured output.

\begin{quote}
\textit{Template:} “Given the Detail Crop image and the Semantic Spotlight image, identify the following attribute keys for the target object: \{key\_set\}. Provide reasoning steps before outputting a JSON object.”
\end{quote}

The final output is constrained to a JSON structure containing the predicted attribute values.

\subsubsection{LLM-as-Judge Scoring Prompt}

For the Continuous Cross-Task Agreement (CCTA) evaluation metric, we employ an LLM-as-Judge paradigm to assess semantic agreement between model predictions and ground-truth answers. Instead of relying solely on exact string matching, the judge evaluates whether the predicted answer conveys the same semantic meaning as the reference answer. The evaluation prompt is structured as follows:

\begin{quote}
“Compare the ground-truth answer: \{gt\_answer\} with the predicted answer: \{pred\_answer\}. Assign a score from 0 to 5 based on semantic equivalence. 0 means fully wrong, 5 means fully correct.”
\end{quote}

Scores are interpreted as a graded measure of semantic similarity, where 0 indicates no semantic overlap and 5 represents full semantic equivalence. This scoring mechanism enables a more nuanced evaluation of model outputs, particularly for open-ended responses where multiple valid phrasings may exist.

\subsection{Human Validation Studies}
\label{sec:human_studies}

To verify the reliability of the proposed benchmark pipeline, we conducted three independent human evaluation studies involving five Master's-level computer science students. These studies assess the correctness and robustness of three key components of the benchmark: (1) the quality of the generated scene graphs, (2) the fidelity of the prompt generation process, and (3) the reliability of the automated LLM-as-Judge scoring procedure. To ensure consistent and interpretable annotations, we developed a dedicated web-based evaluation interface. We also report Krippendorff's alpha ($\alpha$), which measures inter-annotator reliability while accounting for chance agreement. The resulting agreement scores for the three evaluation tasks are reported in \cref{tab:krippendorff}.

The observed agreement levels indicate substantial consistency among annotators across all tasks. In particular, the LLM-as-Judge reliability study achieves the highest agreement ($\alpha = 0.84$), suggesting that human evaluators largely concur with the scoring decisions produced by the automated evaluation framework. The scene graph and prompt fidelity evaluations also demonstrate strong agreement, confirming the robustness of both the scene graph extraction pipeline and the prompt generation procedure.

\begin{table}[ht]
\centering
\caption{Inter-annotator agreement measuring the reliability of the three human evaluation tasks.}
\label{tab:krippendorff}
\begin{tabular}{lc}
\toprule
\textbf{Evaluation Task} & \textbf{Krippendorff's $\alpha$} \\
\midrule
Scene Graph Quality & 0.78 \\
Prompt Fidelity & 0.75 \\
LLM-as-Judge Reliability & 0.84 \\
\bottomrule
\end{tabular}
\end{table}

\subsubsection{Study 1: Scene Graph Quality Evaluation}

\begin{figure}[!ht]
    \centering
    \begin{subfigure}{\textwidth}
        \centering
        \includegraphics[width=0.9\linewidth]{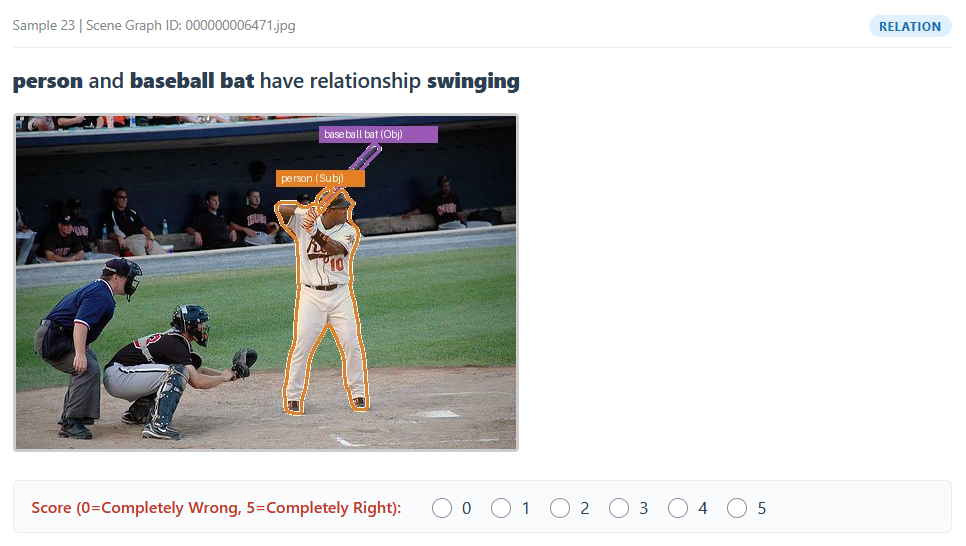}
        \caption{Relational Quality: Annotators verify the extracted relations against the image.}
    \end{subfigure}
    \vspace{1em}
    \begin{subfigure}{0.48\textwidth}
        \centering
        \includegraphics[width=\linewidth]{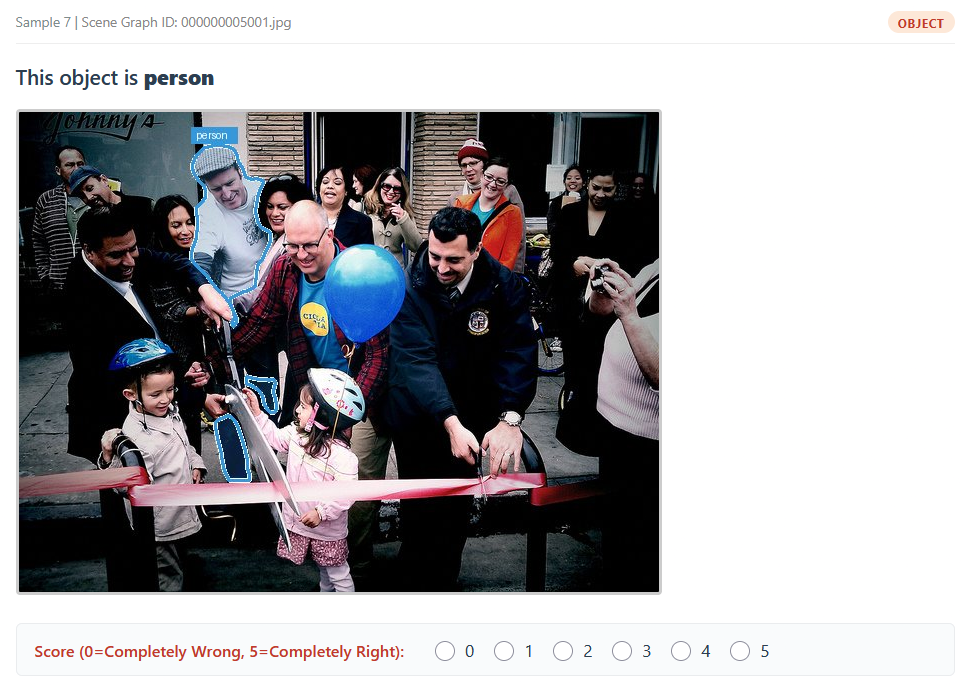}
        \caption{Object Quality: Verification of object existence and segmentation bounds.}
    \end{subfigure}
    \hfill
    \begin{subfigure}{0.48\textwidth}
        \centering
        \includegraphics[width=\linewidth]{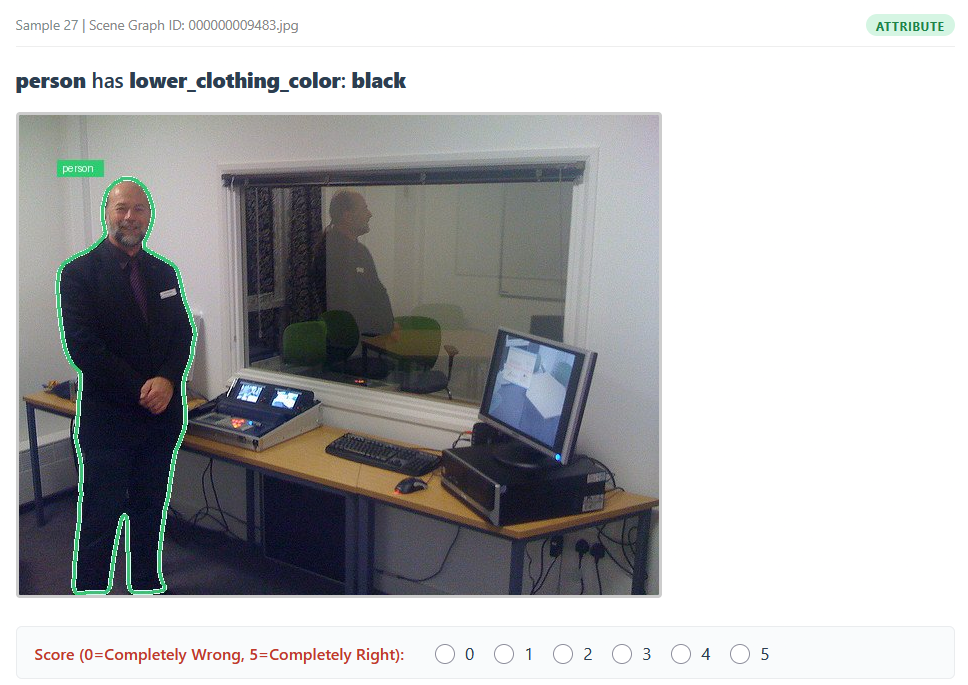}
        \caption{Attribute Quality: Verification of visually grounded attributes.}
    \end{subfigure}
    \caption{The web-based platforms used for the Scene Graph Quality Evaluation. Annotators assess the correctness of extracted objects, attributes, and relations using segmentation mask overlays.}
    \label{fig:sg_quality_interface}
\end{figure}

In the first study, annotators reviewed the extracted scene graph components: objects, attributes, and relations. The interface (\cref{fig:sg_quality_interface}) displayed segmentation masks next to the corresponding text. Annotators rated each fact on a 0 to 5 scale to judge how well it matched the source image. 

The annotators showed strong agreement with each other ($\alpha=0.78$). The evaluation achieved an overall accuracy of $94.02\%$. Relational facts proved to be the most reliable, achieving $95.17\%$ accuracy.

\subsubsection{Study 2: Prompt Fidelity Evaluation}

\begin{figure}[!ht]
    \centering
    \begin{subfigure}{\textwidth}
        \centering
        \includegraphics[width=0.9\linewidth]{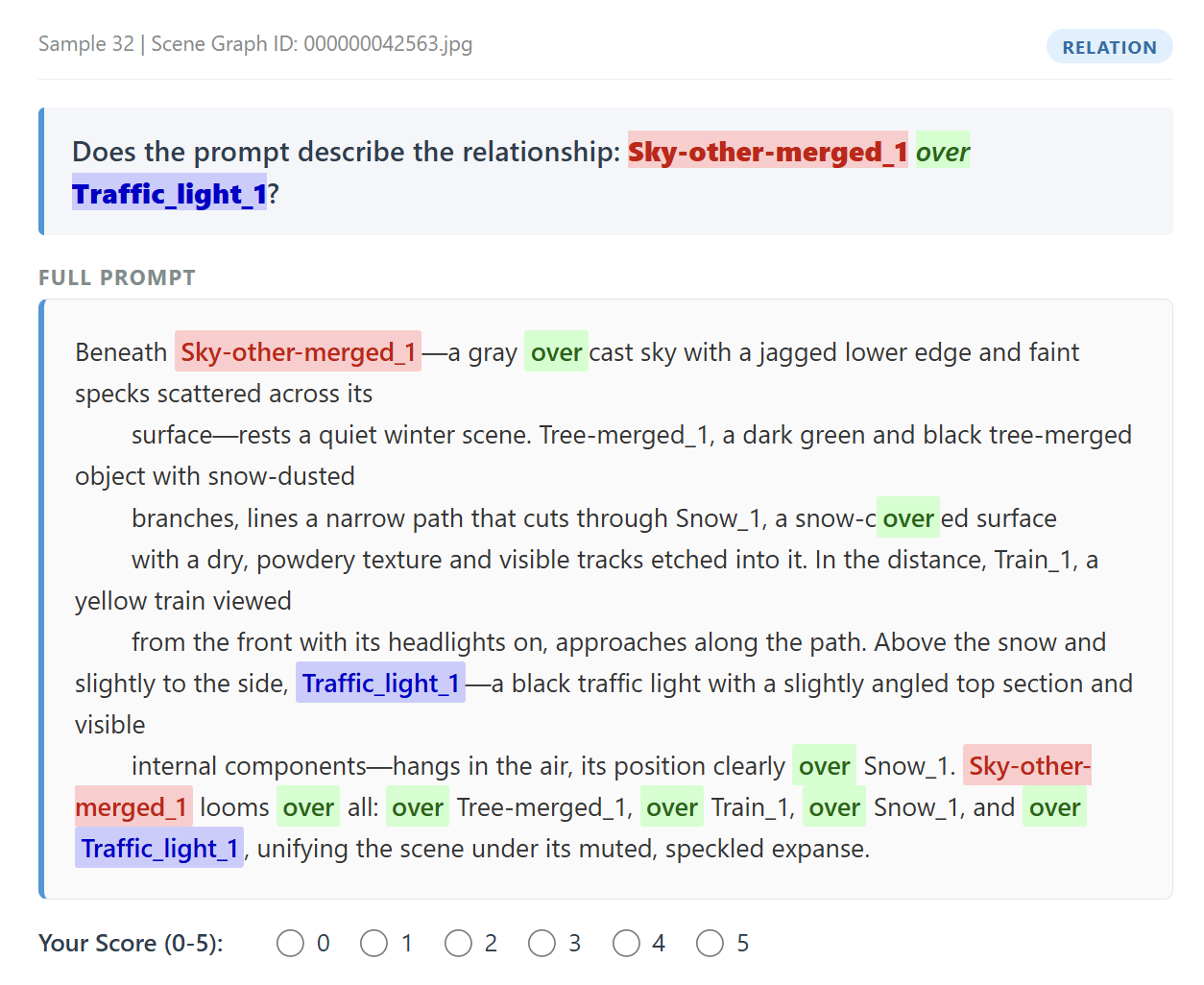}
        \caption{Relational Interaction: \textbf{\color{red}Subject}, \textbf{\color{blue}Object}, and \textbf{\color{green}Predicate} highlights.}
    \end{subfigure}
    \vspace{1em}
    \begin{subfigure}{0.48\textwidth}
        \centering
        \includegraphics[width=\linewidth]{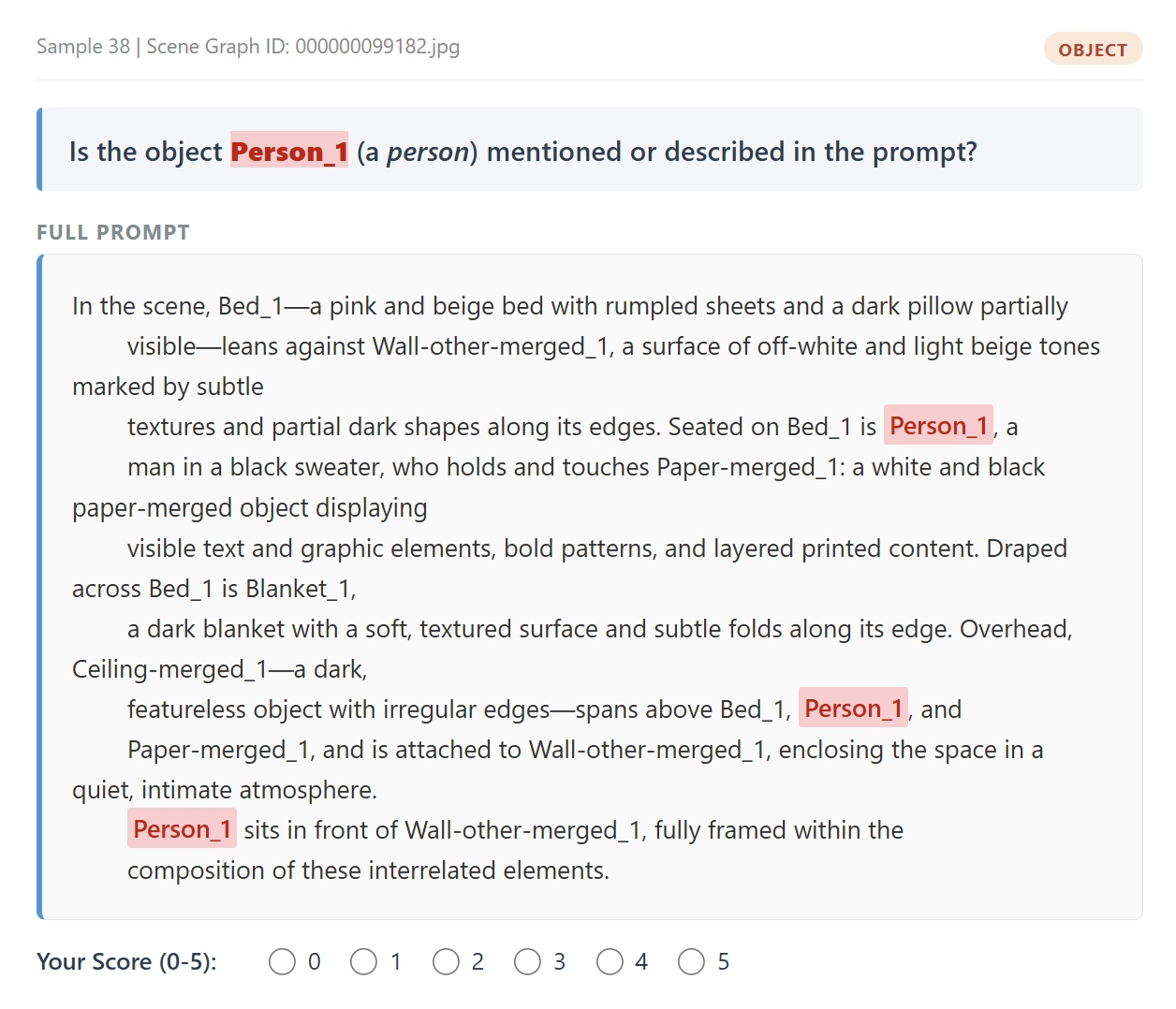}
        \caption{Object Existence: \textbf{\color{red}Target Object} identification.}
    \end{subfigure}
    \hfill
    \begin{subfigure}{0.48\textwidth}
        \centering
        \includegraphics[width=\linewidth]{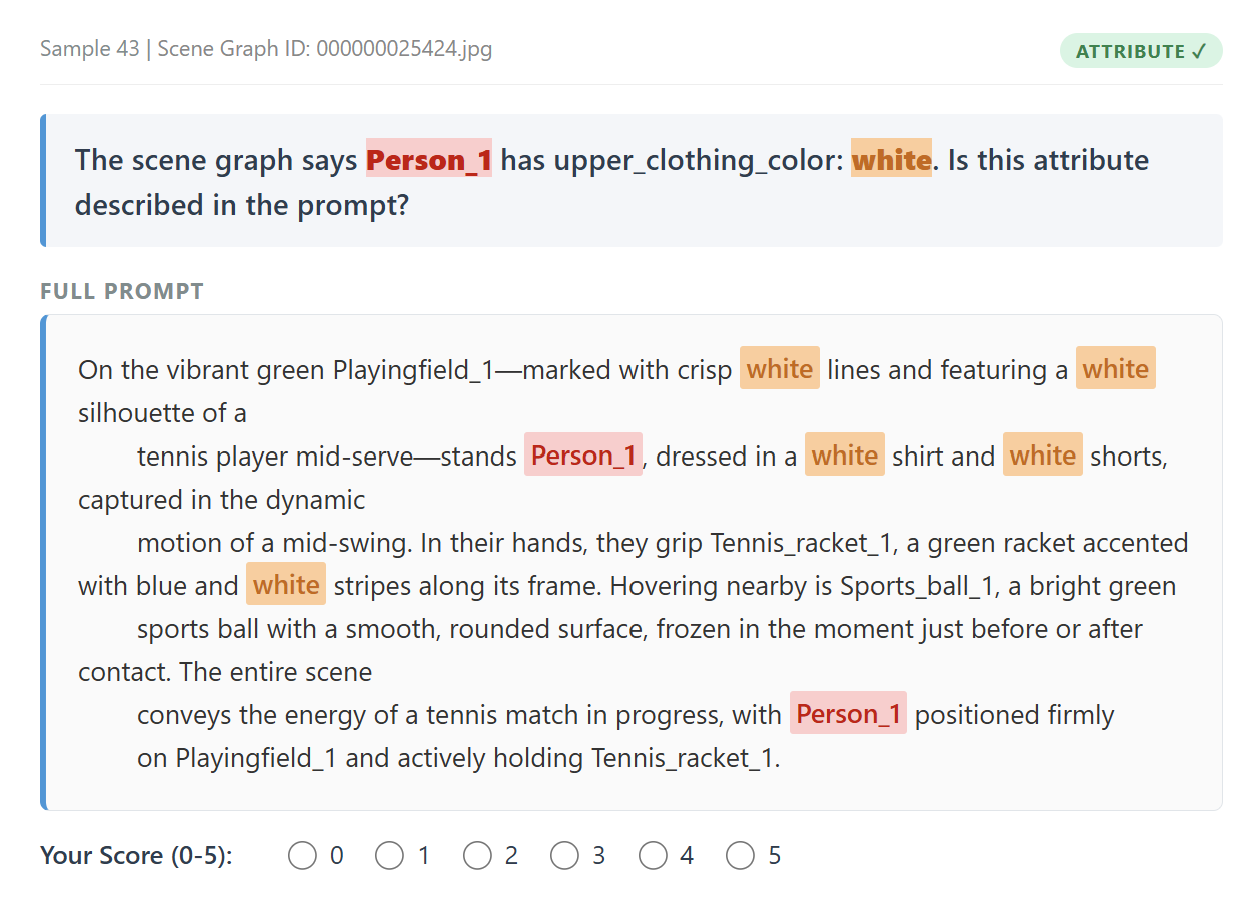}
        \caption{Attribute Description: \textbf{\color{orange}Fact-Specific Attribute} highlights.}
    \end{subfigure}
    \caption{The interface for prompt fidelity evaluation. Color-coded highlighting links atomic facts within natural language prompts to their image locations.}
    \label{fig:human_interface}
\end{figure}

The second study checked whether the scene graph linearization and prompt refinement accurately translated the scene graph's facts into natural language. We used a similar interface for this task (\cref{fig:human_interface}). We color-coded the highlights so annotators could easily map specific facts from the dense prompt text to the visual evidence in the image.

The annotators showed strong agreement with each other ($\alpha=0.75$). The results confirmed the generated prompts preserved the original scene graph information. The language model rarely hallucinated or dropped details during linearization.

\subsubsection{Study 3: LLM-as-Judge Reliability}

\begin{figure}[!ht]
    \centering
    \begin{subfigure}{0.48\textwidth}
        \centering
        \includegraphics[width=\linewidth]{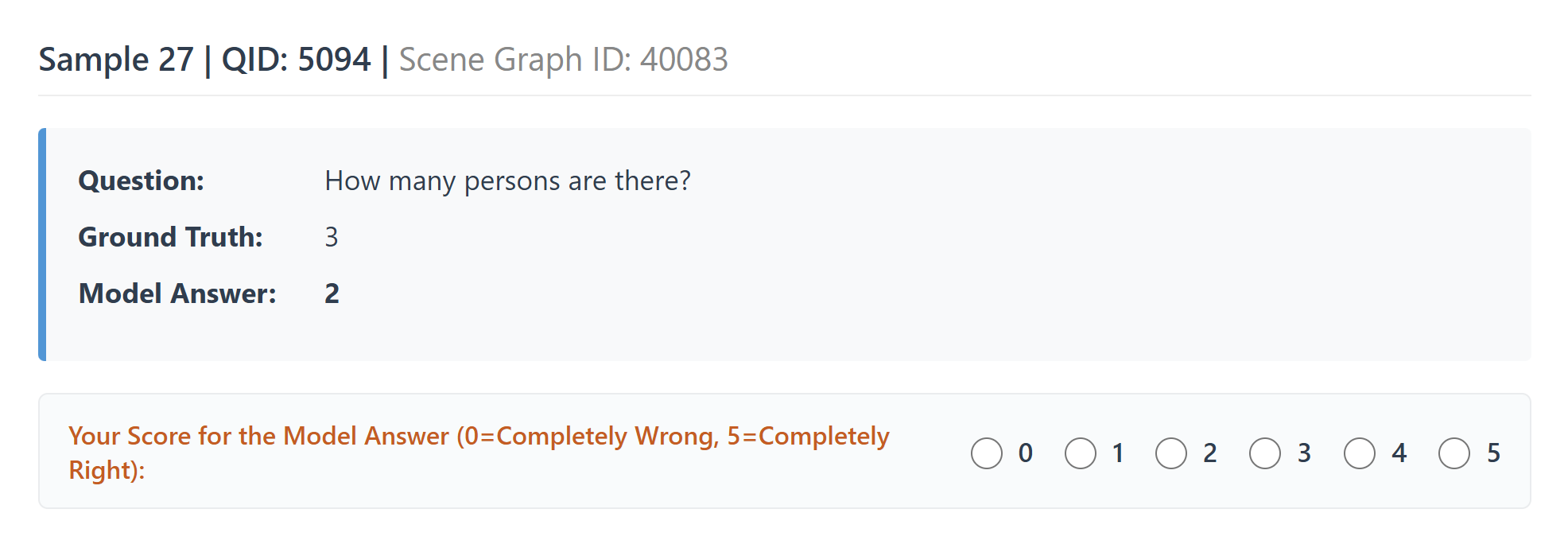}
        \caption{Counting: Judge compares GT (3) vs. Model (2).}
    \end{subfigure}
    \hfill
    \begin{subfigure}{0.48\textwidth}
        \centering
        \includegraphics[width=\linewidth]{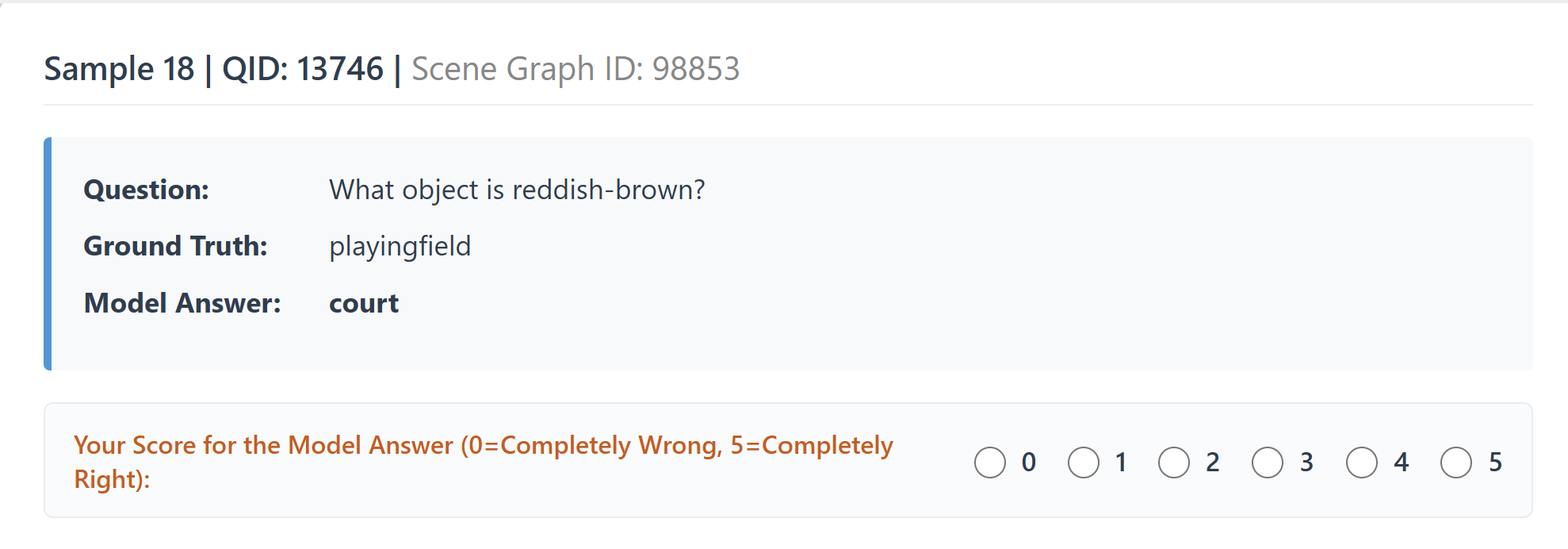}
        \caption{Labeling: Verifying "playingfield" vs "court".}
    \end{subfigure}
    \caption{The interface for LLM-Judge Reliability. Annotators score semantic equivalence of Model Answers against Ground Truth on a 0--5 scale.}
    \label{fig:judge_interface}
\end{figure}

The final study evaluated our automated scorer. Annotators used the interface (\cref{fig:judge_interface}) to rate the semantic equivalence between the model's answers and the ground truth on a 0 to 5 scale. They looked at object counting, label identification, and attribute or relational reasoning. 

The annotators showed strong agreement with each other ($\alpha=0.84$). We then compared these human ratings to the LLM's scores. As Table 4 in the main paper shows, the Pearson correlation scores are high ($r=0.800$ for attributes and $r=0.850$ for relationships). The automated judge ranks quality in a way that matches human intuition.

\subsection{Additional Analysis and Qualitative Results}

\subsubsection{Performance Across Attribute and Relation Categories}

To complement the aggregate benchmark results, we provide a detailed analysis of model performance across fine-grained attribute and relation dimensions. For each evaluated model, we visualize generation and understanding performance separately to highlight potential asymmetries between these two capabilities.

In the visualizations, the left side of each plot represents \textit{generation} performance, while the right side represents \textit{understanding} performance. A central bar indicates the imbalance between the two scores, identifying whether a model leans toward content creation or interpretation. Attributes are aggregated into six semantic dimensions: \emph{Color \& Material}, \emph{Environment}, \emph{State \& Functionality}, \emph{Type \& Parts}, \emph{Text, Symbols \& Counts}, and \emph{Pose, View \& Placement}. For relational reasoning, we utilize the taxonomy introduced in the previous section, evaluating performance across the \emph{Locomotion}, \emph{Posture}, \emph{Social}, and \emph{Spatial} categories. This approach reduces noise from individual predicates and highlights broader reasoning trends across different model architectures.

\begin{figure*}[!ht]
    \centering
    \begin{subfigure}{0.95\textwidth}
        \centering
        \includegraphics[width=\linewidth]{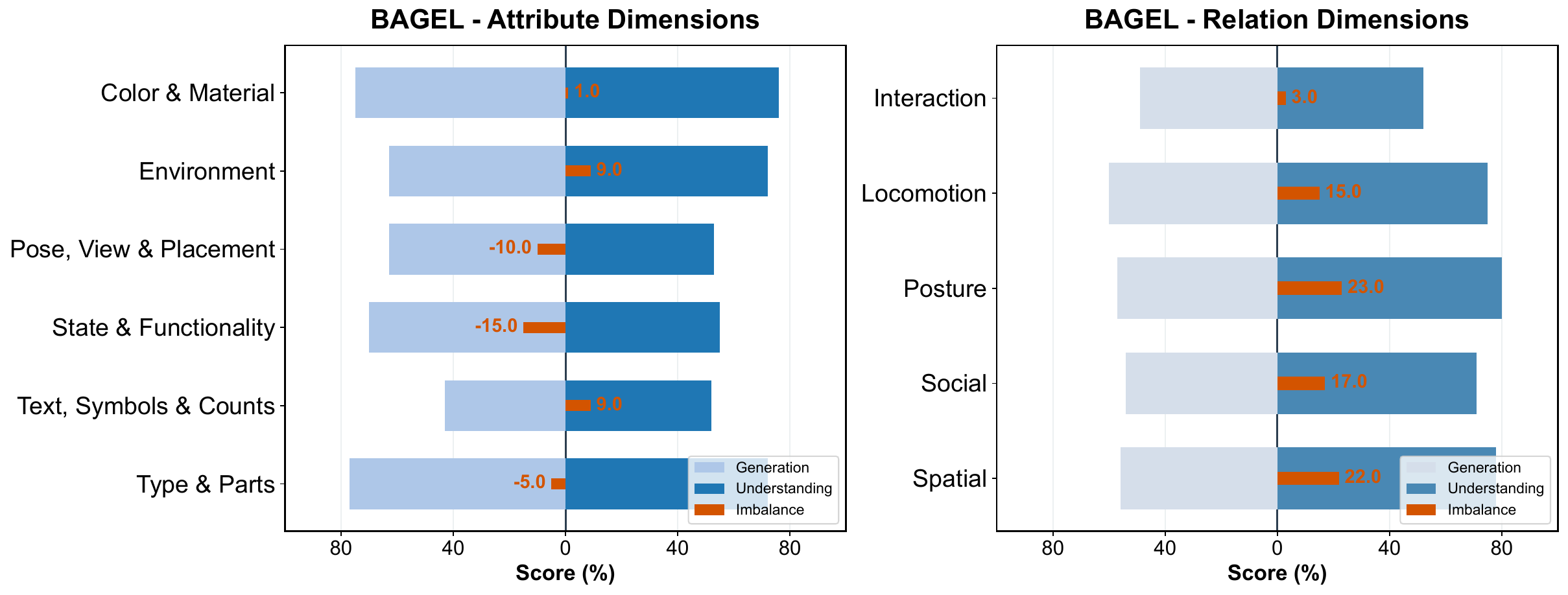}
        \caption{BAGEL.}
    \end{subfigure}
    
    \vspace{1em}
    
    \begin{subfigure}{0.95\textwidth}
        \centering
        \includegraphics[width=\linewidth]{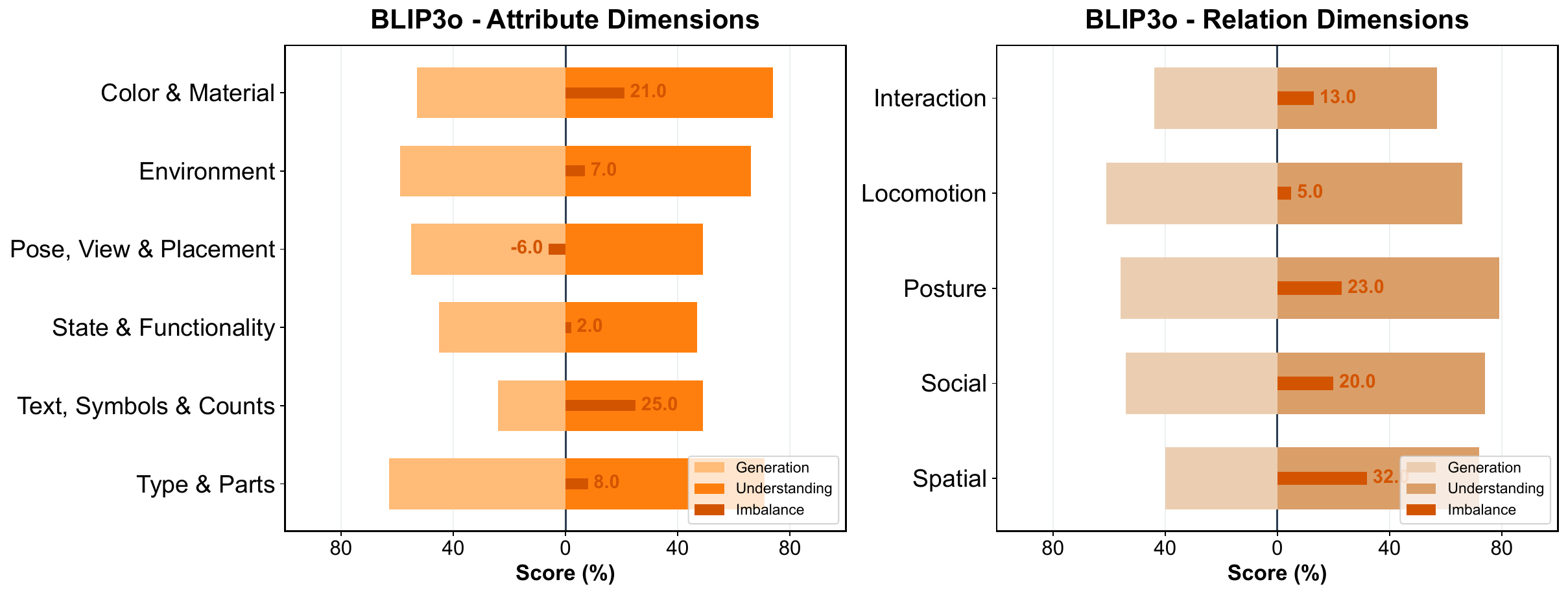}
        \caption{BLIP3o.}
    \end{subfigure}
    \caption{Per-dimension tornado plots for BAGEL and BLIP3o. Each plot shows attribute and relation category scores for generation and understanding, along with the corresponding imbalance.}
    \label{fig:tornado_bagel_blip3o}
\end{figure*}

\begin{figure*}[!ht]
    \centering
    \begin{subfigure}{0.95\textwidth}
        \centering
        \includegraphics[width=\linewidth]{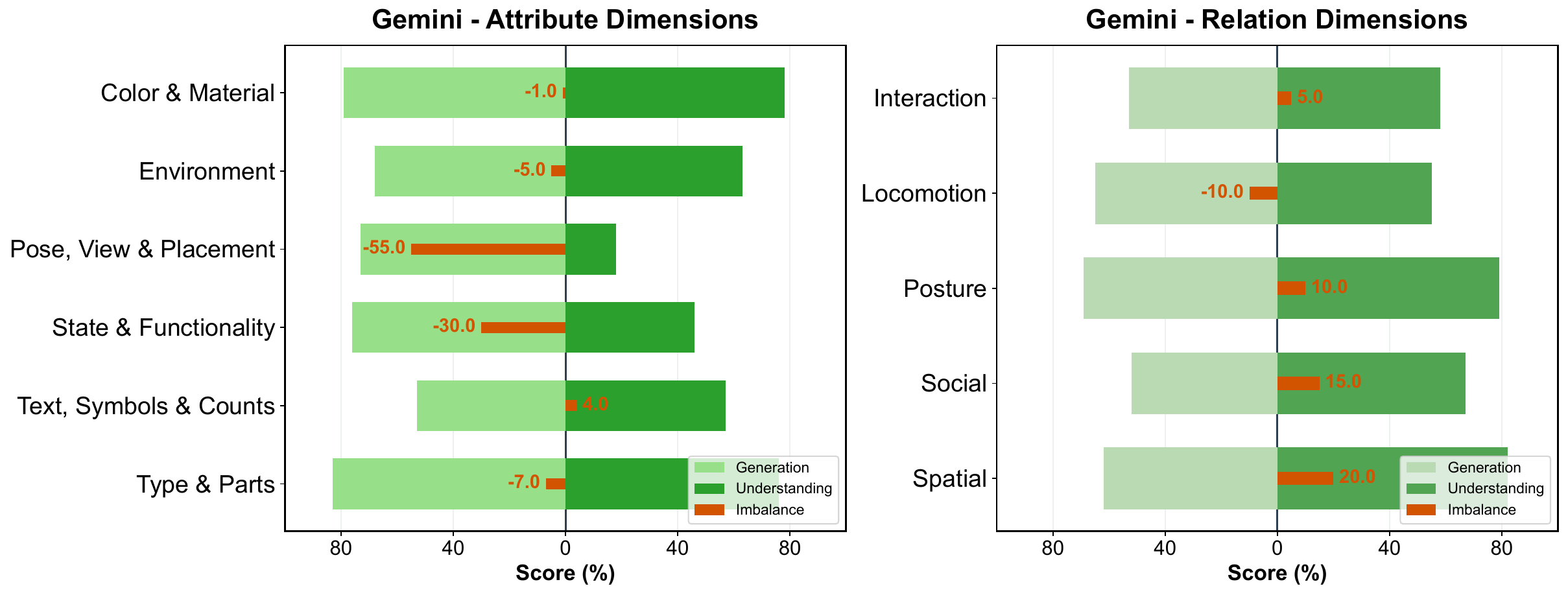}
        \caption{Gemini.}
    \end{subfigure}
    
    \vspace{1em}
    
    \begin{subfigure}{0.95\textwidth}
        \centering
        \includegraphics[width=\linewidth]{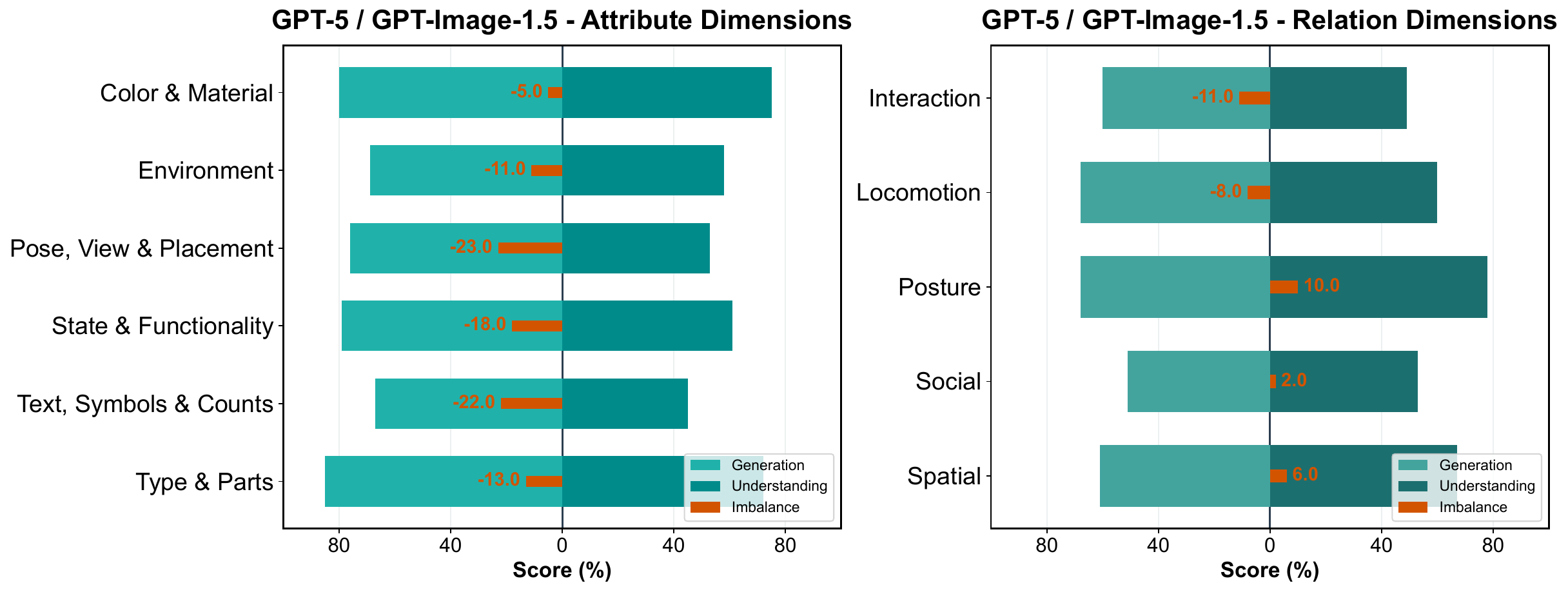}
        \caption{GPT.}
    \end{subfigure}
    \caption{Per-dimension tornado plots for Gemini and GPT.}
    \label{fig:tornado_gemini_gpt}
\end{figure*}

\begin{figure*}[!ht]
    \centering
    \begin{subfigure}{0.95\textwidth}
        \centering
        \includegraphics[width=\linewidth]{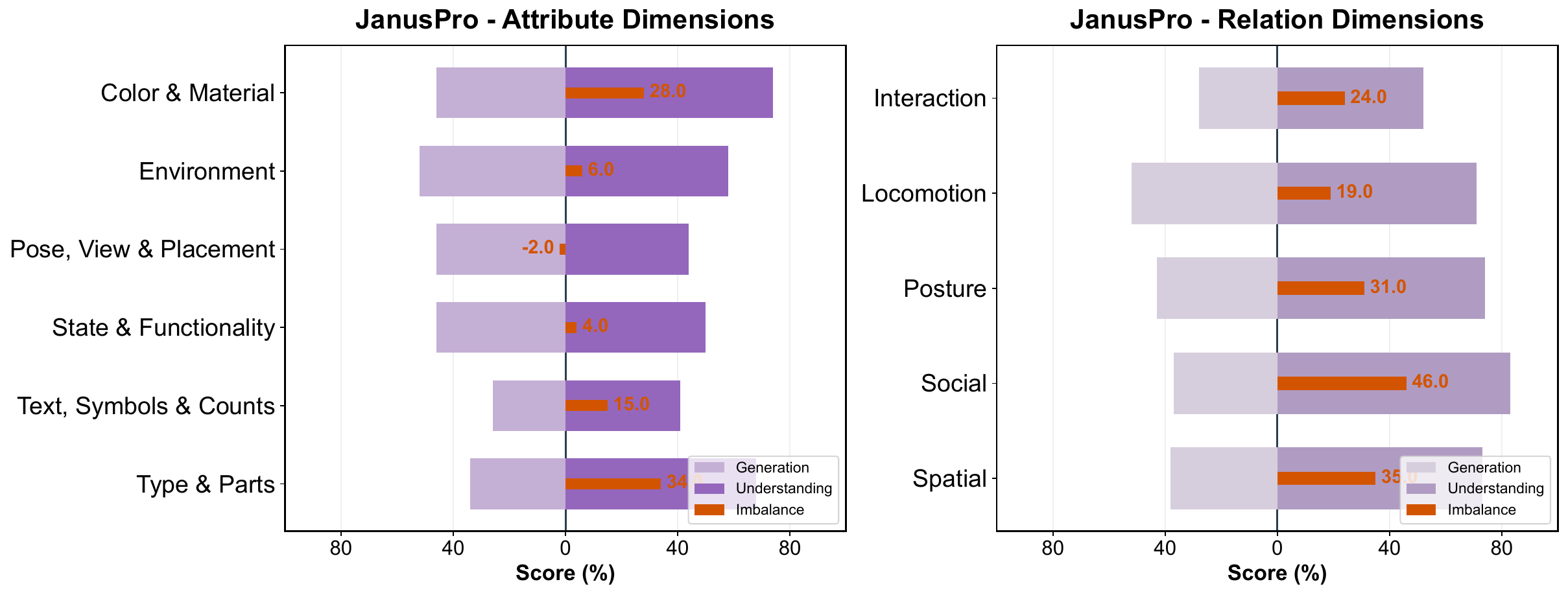}
        \caption{JanusPro.}
    \end{subfigure}
    
    \vspace{1em}
    
    \begin{subfigure}{0.95\textwidth}
        \centering
        \includegraphics[width=\linewidth]{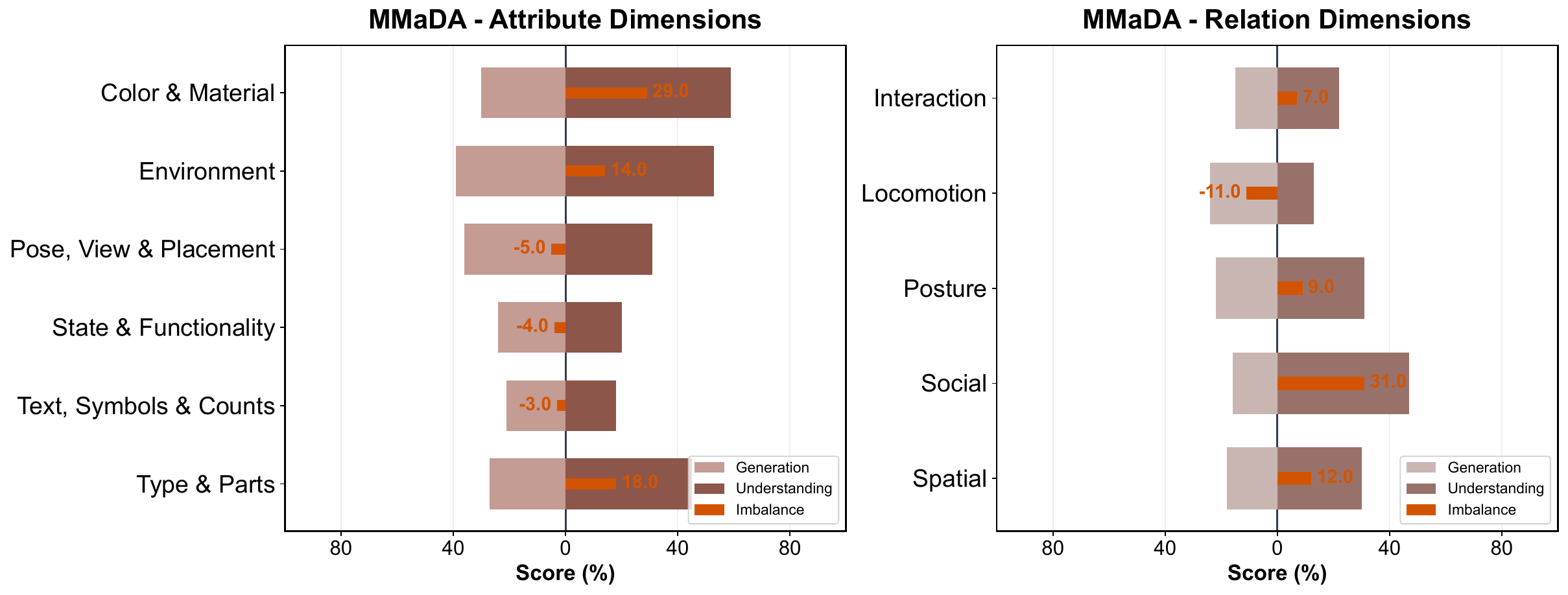}
        \caption{MMaDA.}
    \end{subfigure}
    \caption{Per-dimension tornado plots for JanusPro and MMaDA.}
    \label{fig:tornado_januspro_mmada}
\end{figure*}

\begin{figure*}[!ht]
    \centering
    \begin{subfigure}{0.95\textwidth}
        \centering
        \includegraphics[width=\linewidth]{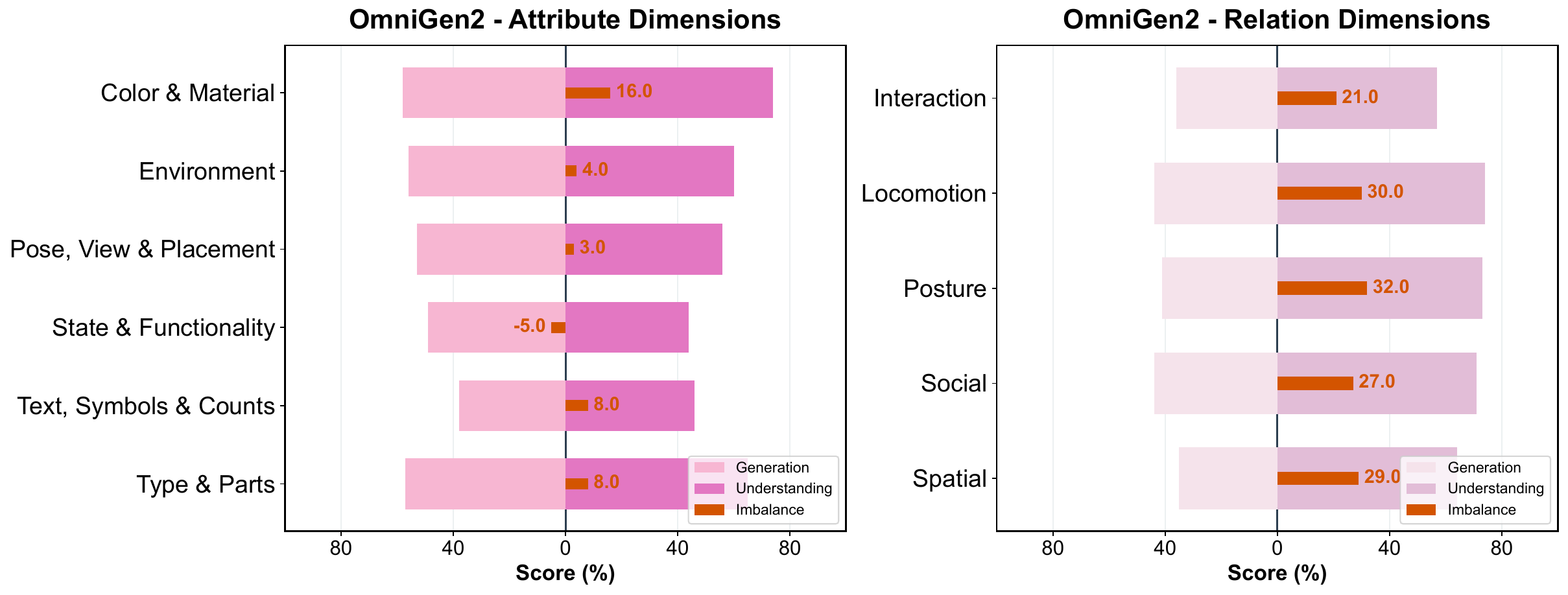}
        \caption{OmniGen2.}
    \end{subfigure}
    
    \vspace{1em}
    
    \begin{subfigure}{0.95\textwidth}
        \centering
        \includegraphics[width=\linewidth]{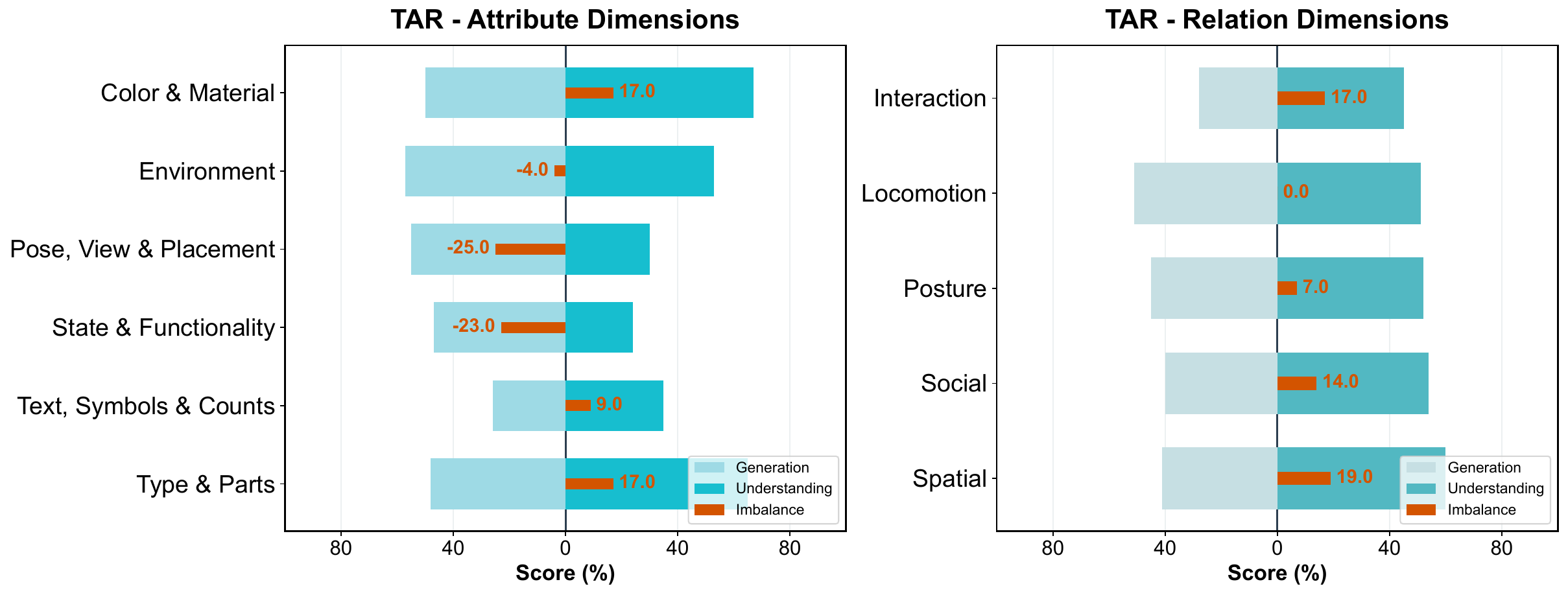}
        \caption{Tar.}
    \end{subfigure}
    \caption{Per-dimension tornado plots for OmniGen2 and Tar}
    \label{fig:tornado_omnigen2_showo}
\end{figure*}

\begin{figure*}[!ht]
    \centering
    \begin{subfigure}{0.95\textwidth}
        \centering
        \includegraphics[width=\linewidth]{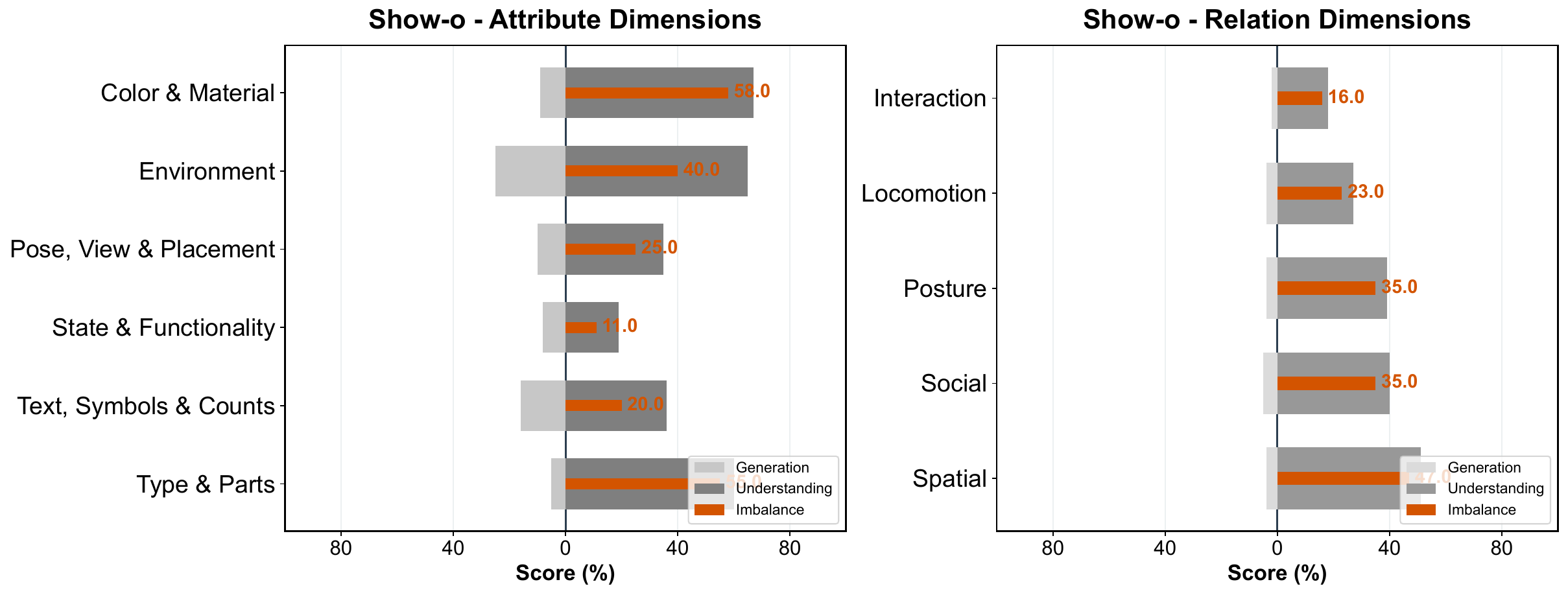}
        \caption{Show-o.}
    \end{subfigure}
    
    \vspace{1em}
    
    \begin{subfigure}{0.95\textwidth}
        \centering
        \includegraphics[width=\linewidth]{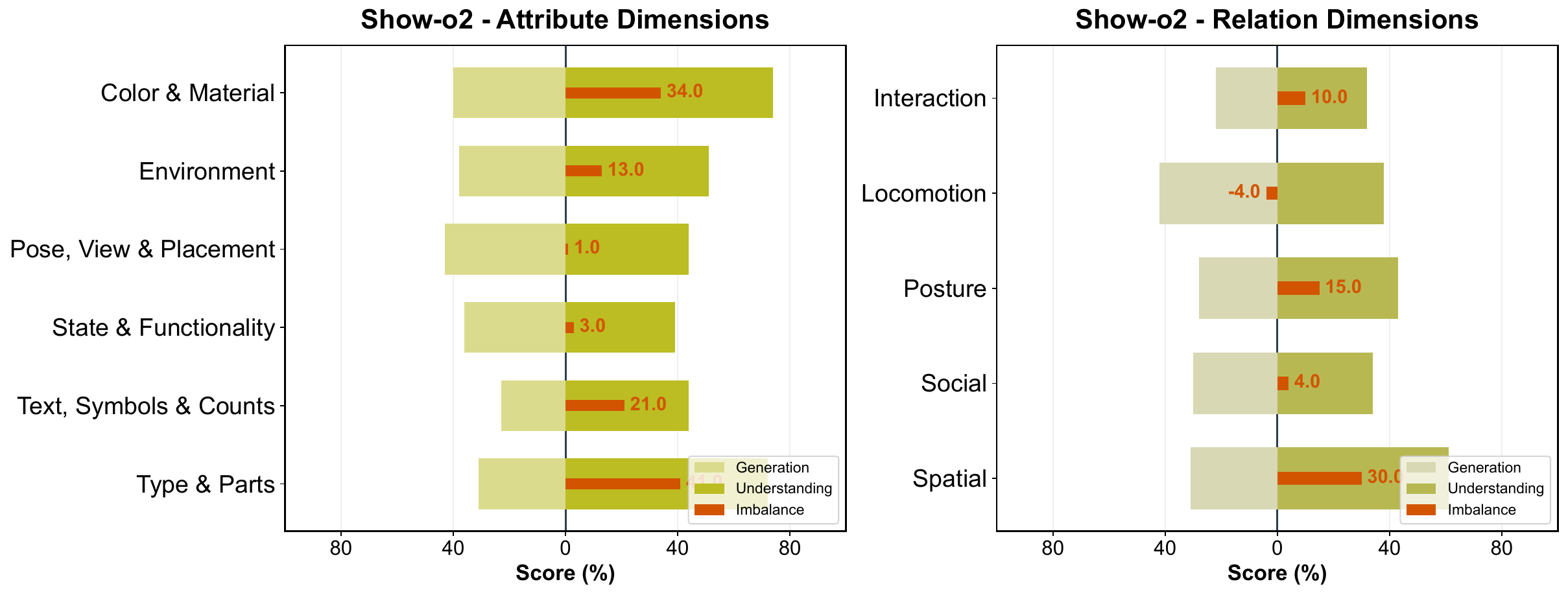}
        \caption{Show-o2.}
    \end{subfigure}
    \caption{Per-dimension tornado plots for Show-o and Show-o2.}
    \label{fig:tornado_showo2_tar}
\end{figure*}

As shown in~\cref{fig:tornado_bagel_blip3o,fig:tornado_gemini_gpt,fig:tornado_januspro_mmada,fig:tornado_omnigen2_showo,fig:tornado_showo2_tar}, we observe a consistent trend in which models achieve higher scores on \textit{understanding} tasks compared to \textit{generation}. This pattern suggests that interpreting existing relational structures is generally easier for current multimodal models than producing accurate relational descriptions during generation.

A similar but more nuanced pattern emerges for attribute reasoning. Weaker open-source models such as \emph{Show-o}, \emph{Show-o2}, \emph{OmniGen2}, \emph{MMaDA}, and \emph{JanusPro} demonstrate a consistent advantage in understanding across nearly all attribute dimensions. These models appear to rely heavily on recognition capabilities while struggling to reliably generate attribute-consistent outputs.

Stronger open-source models, including \emph{TAR} and \emph{BAGEL}, display a more heterogeneous pattern. For some attribute dimensions—such as \emph{Color \& Material} and \emph{Pose, View \& Placement}—understanding remains dominant, while, in other dimensions, generation performance approaches or occasionally surpasses understanding.

Overall, these qualitative analyses reveal a systematic asymmetry between generation and understanding capabilities across models, highlighting an important challenge for future multimodal systems: achieving balanced performance across both tasks rather than excelling primarily in one direction.

\clearpage

\end{document}